\documentclass{article}

\usepackage{microtype}
\usepackage{graphicx}
\usepackage{subfigure}
\usepackage{booktabs} 
\usepackage{tikz}
\definecolor{sinkcol}{HTML}{E69F00}
\definecolor{windowcol}{HTML}{0072B2}
\definecolor{questcol}{HTML}{009E73}

\newcommand{\sq}[1]{\tikz[baseline=-0.5ex]%
  \fill[#1!80, draw=#1!70!black, rounded corners=0.4pt] (0,-0.7ex) rectangle ++(1.4ex,1.4ex);}

\usepackage{makecell}
\usepackage{color,soul}
\definecolor{myRed}{rgb}{0.60,0.10,0.15}
\definecolor{myGreen}{rgb}{0.25,0.60,0.15}
\definecolor{myBlue}{rgb}{0.15,0.30,0.60}
\definecolor{myOrange}{rgb}{0.70,0.50,0.25}
\newcommand{\red}[1]{{\textbf{\color{myRed}#1}}} 
\newcommand{\green}[1]{{\textbf{\color{myGreen}#1}}}

\newcommand{\orange}[1]{{\textbf{\color{myOrange}#1}}}

\usepackage{listings}
\usepackage{amsmath}
\usepackage{amssymb}

\usepackage{colortbl}
\definecolor{cellRed}{rgb}{0.97,0.78,0.78}
\definecolor{cellGreen}{rgb}{0.80,0.93,0.78}

\usepackage{rotating}
\usepackage{placeins}
\usepackage{hyperref}

\usepackage[preprint]{mlsys2026}

\mlsystitlerunning{Accelerating Long-Context Decoding in Large-Scale Hybrid LLMs Without Retraining}

\begin{document}

\twocolumn[
\mlsystitle{Accelerating Long-Context Decoding in Large-Scale Hybrid LLMs Without Retraining}
\mlsyssetsymbol{equal}{*}
\begin{mlsysauthorlist}
\mlsysauthor{Jo\~ao E. Batista}{rccs}
\mlsysauthor{Emil Vatai}{rccs}
\mlsysauthor{Mohamed Wahib}{rccs}
\end{mlsysauthorlist}
\mlsysaffiliation{rccs}{RIKEN Center for Computational Science, Kobe, Hyogo, Japan}
\mlsyscorrespondingauthor{Jo\~ao E. Batista}{joao.batista@riken.jp}
%
\mlsyskeywords{Machine Learning, MLSys}
\vskip 0.3in
\begin{abstract} 
The full-attention layers of large language models (LLMs) make every generated token more expensive than the previous one, slowing models down on long-context tasks. 
Hybrid models reduce this cost by replacing most full-attention layers with linear-attention layers, but the first still dominate decoding at long contexts. 
We accelerate these remaining layers without any training or weight changes using combined sparse attention: each layer attends to a sliding window of recent tokens, a pinned prefix, and a fixed budget of most relevant tokens in the context.
The window size and budget are found on a per-layer basis using a multi-objective search algorithm.
Across six hybrid models from three families, from 4B to 397B parameters, the selected configurations preserve their baseline model's accuracy on 21 long-context datasets. 
Decoding 64k/256k tokens becomes up to 1.26x/2.22x faster on Qwen3.5-4B, 1.20x/2.23x on Qwen3.5-122B-A10B, and 1.16x/2.06x on Ornith-1.0-397B.
\end{abstract}
]

\printAffiliationsAndNotice{}  

\section{Introduction}
\label{sec:intro}
\begin{figure}[t]
    \centering
    \includegraphics[width=.95\linewidth]{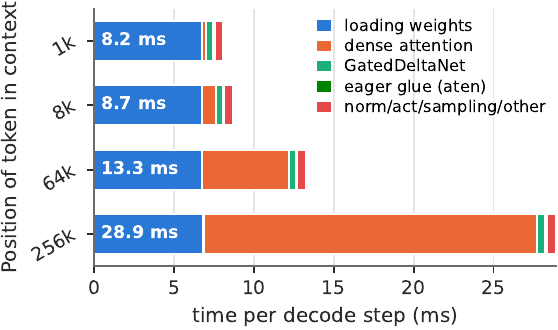}
    \vspace{-2mm}
    \caption{Time spent decoding one token by Qwen3.5-4B in each kernel (averaging 256 tokens), at different context lengths.}
    \label{fig:profile_context}
\end{figure}

Current implementations of self-attention (flash-attention~\cite{flash}) cost $O(L^2)$ time and $O(L)$ memory in the sequence length $L$. For autoregressive inference, the memory capacity is usually the binding constraint: the KV cache grows linearly with every generated token and with every concurrent request. At long context, the full-attention layers consume the majority of the decode-step budget and determine how many concurrent requests a deployment can serve (Figure~\ref{fig:profile_context}).

\par
Two lines of work have emerged to defuse this bottleneck. The first is architectural: replace softmax attention with a recurrent, convolutional, or windowed mixer whose per-step cost is constant in $L$. Mamba~\cite{mamba}, RWKV~\cite{rwkv}, Hyena~\cite{hyena}, Gated DeltaNet~\cite{7}, and the sliding-window layers of Gemma~4~\cite{gemma4} belong here, together with the \emph{hybrid} models (Qwen3.5~\cite{11}, Gemma~4, gpt-oss~\cite{gptoss}, MiniMax-Text-01~\cite{minimax}, Kimi-K3~\cite{kimik3}) that interleave a minority of full-attention layers with cheaper attention layers. The accuracy of hybrids is competitive, but decisions such as how many full-attention layers to use, where to place them, and their window size are frozen into the released weights at pretraining time.

\par
The second line is post-training linearization: take a pretrained Transformer, swap its attention for a linear variant, and recover the accuracy with continued training. Most works focus on smaller models, up to 8B parameters; LolCats~\cite{lolcats} is one exception, applied to Llama 3.1 405B. Besides focusing on smaller models, these works often require large amounts of training data, with SUPRA~\cite{supra} reporting 5\% of pretraining, and LawCat~\cite{lawcat} reporting 0.1\% of pretraining. LolCats use a model with size comparable to our work, and report spending 454 H100 GPU-hours in attention transfer and LoRA fine-tuning. Although Jet-Nemotron-2B obtained a 12.0x speedup over its base model at a 64k-token context, it was trained on 600B tokens, using 18,328 H100 GPU-hours~\cite{jetnemo}. It is not practical to expend this compute and data cost for every new released model, motivating the search for cheaper alternatives.

\par
This work builds upon hybrid models whose remaining full-attention layers still carry the long-context cost, and reduces their computational cost without further training by limiting their attention. Qwen3.5-27B keeps 16 full-attention layers out of 64; gpt-oss-120b, 18 out of 36; Ornith-397B, 15 out of 60. These layers read $L$ keys per step, and they are the ones that would benefit from optimizations. The question that motivates this work is therefore:

\par
\textit{Given a pre-trained hybrid model, how much can we restrict the attention of its full-attention layers at no accuracy degradation, without retraining and without editing weights?}

\par
Attention maps are known to be highly concentrated, making training-free KV eviction (H2O~\cite{h2o}, DuoAttention~\cite{duoattention}) work, but this is an empirical claim, and recent large-scale evaluation~\cite{6} shows it is strongly task- and scale-dependent. We hypothesize that, since attention is sparse at long context, each query only needs the keys of a small, query-dependent subset of positions. By reducing attention in models, we obtain speedups by skipping the bottleneck of decoding in long-context tasks, full attention (Figure \ref{fig:profile_context}). This motivates searching for configurations that minimize the cost of attention through linearization techniques, while aiming to preserve the accuracy of the baseline models.

\par
Our challenge is therefore twofold: first, identifying \emph{which} positions carry information depending on the query and the layer, not only on the position; second, minimizing the attention budget on a layer-by-layer basis. We combine a plain sliding window ~\cite{1,2,3} with a dynamic query-aware top-$k$ selection on paged KV (Quest ~\cite{quest}) on a layer-by-layer basis, and let a multi-objective search decide the \textit{window} and \textit{budget} of every layer.

\par \textbf{Contributions:}
\begin{itemize}
    \item A per-layer, training-free sparse-attention configuration search algorithm that combines a sliding window, a pinned prefix, and query-aware paged selection, installed into vLLM~\cite{8}. 
    \item A multi-objective NSGA-II~\cite{nsga2} search heuristic that returns per-layer (window, budget) pairs that optimize both long-context accuracy and estimated decoding cost. The search costs 8.6--34.2 GPU-hours on A100 GPUs for models ranging from 4 to 120B parameters, and 31.3--58.4 GPU-hours on a x4 GB200 GPU node for models with 122B and 397B parameters.
    \item An evaluation on six hybrid models from three families (Qwen3.5 at 4B / 27B / 122B, Ornith 397B, Gemma-4-31B, gpt-oss-120b) across 21 long-context benchmarks. To our knowledge, this is the first training-free sparse-attention method evaluated at the 100B+ scale.
\end{itemize}

\par \textbf{Headline results.} On every model, at least one configuration preserves the baseline accuracy on every long-context dataset, with averages within 0.5 percentage points (pp) of the baseline. At 64k/256k tokens and batch size 4, these configurations decode 1.26/2.22$\times$ faster on Qwen3.5-4B, 1.20/2.23$\times$ on Qwen3.5-122B-A10B, and 1.16/2.06$\times$ on Ornith-1.0-397B. The gains are smaller in the other models: on the dense Qwen3.5-27B (1.11$\times$ at 64k tokens) and Gemma-4-31B (1.22$\times$ at 128k tokens) models, where a larger chunk of the decoding time is used to read weights (Figure~\ref{fig:profile_decode}, in appendix), and on gpt-oss-120b (1.16$\times$ at 128k tokens), where our method preserved most of the full-attention layers. At short context, every configuration is slower than the unmodified model.

\section{Related Work}
\label{sec:related}

Reducing the cost of self-attention in deployed language models is studied extensively. We group prior work by the computational requirements: full pretraining, a post-training phase of continued pretraining, or no training at all. Table~\ref{tab:related} summarizes the landscape and locates this work in it.

\begin{table*}[!t]
\centering
\caption{Methods for reducing attention costs in LLMs. The first block covers pretraining methods, the second covers post-training, and the third is training-free. Our pipeline is training-free and tested in hybrid models across three model families (Qwen3.5, Gemma~4, gpt-oss). \green{Green}: no training, long-context capability, or more than 300B parameters; \orange{orange}: post-training, limited long-context capability, or 10--300B parameters; \red{red}: full pretraining or over 1B token training, no long-context evaluation, or fewer than 10B parameters. }
\footnotesize
\setlength{\tabcolsep}{3pt}
\renewcommand{\arraystretch}{0.87}
\begin{tabular}{>{\raggedright\arraybackslash}p{2.7cm} >{\raggedright\arraybackslash}p{3.35cm} >{\raggedright\arraybackslash}p{3.35cm} >{\raggedright\arraybackslash}p{4.7cm} >{\raggedright\arraybackslash}p{2.0cm}}
\toprule
\textbf{Method} & \textbf{Approach} & \textbf{Training required} & \textbf{Long context} & \textbf{Largest model reported} \\
\midrule
\multicolumn{5}{l}{\emph{Pretraining -- architecture decided before weights are released}} \\
Mamba, Mamba-3 & SSM  & \red{full pretraining} & \orange{synthetic only (1M), 2k on text} & \red{1.5--2.8B} \\
Hyena & long convolution & \red{full pretraining} & \orange{synthetic recall (131k), 2k on text} & \red{1.3B} \\
RWKV  & linear recurrence & \red{full pretraining} & \orange{16k} & \orange{14B} \\
Qwen3.5, Gemma~4, gpt-oss, kimi-k3 & hybrid full + linear/SWA & \red{full pretraining} & \green{yes (1M)} & \green{2.8T} \\
\midrule
\multicolumn{5}{l}{\emph{Post-training linearization -- continued training on top of a released model}} \\
Hedgehog & learned softmax-like feature map & \orange{LoRA conversion} & \red{not evaluated} & \red{7B} \\
LoLCATs & linear attention swap & \orange{40M tokens, LoRA} & \orange{8k, single dataset} & \green{405B} \\
Liger  & gated recurrence & \orange{20M tokens, LoRA} & \red{not evaluated} & \red{8B} \\
LAWCAT & distillation + convolution& \orange{94M tokens, LoRA} & \orange{up to 22k} & \red{7B} \\
SUPRA  & linear uptraining & \red{20B tokens, uptraining}  & \orange{16k} & \red{7B} \\
HALO  & hybrid distillation & \red{2.3B tokens, distillation}  & \green{yes (256k)} & \red{8B} \\
JetNemotron  & PostNAS + retraining & \red{600B tokens}  & \green{yes (64k)} & \red{4B} \\
\midrule
\multicolumn{5}{l}{\emph{Training-free sparse / windowed attention on frozen weights}} \\
StreamingLLM & window + sinks & \green{none}  & \orange{no retrieval beyond window} & \orange{70B} \\
H2O  & KV eviction & \green{none}  & \orange{2k} & \orange{30B} \\
SnapKV & prefill-time top-$k$ & \green{none}  & \green{yes (380k)} & \orange{47B} \\
Quest & query-aware paged top-$k$ & \green{none}  & \green{yes (100k)} & \red{7B} \\
DuoAttention & retrieval-head split & \orange{gates only, 2k steps} & \green{yes (1M)} & \orange{70B} \\
\midrule
\textbf{Ours}  & per-layer SWA + prefix + Quest + NSGA-II & \green{none}$^{a}$ & \green{yes (21 datasets, up to 64k tokens)} & \green{397B} \\
\bottomrule
\end{tabular}
\par\smallskip
\footnotesize
$^{a}$ No gradient updates to the base model. Configuration search uses up to 60 GPU-hours per model (Table \ref{tab:cost}).
\label{tab:related}
\end{table*}

\subsection{Pretrained Linear and Hybrid Architectures}

Mamba~\cite{mamba}, Hyena~\cite{hyena}, and RWKV~\cite{rwkv} replace softmax attention with a recurrence or a long convolution whose cost grows linearly or quasi-linearly with $L$. They require a full pretraining run, are relatively small (under 15B parameters), and their long-context ability is either shown on synthetic tasks only (Mamba and Hyena, whose language models are trained on contexts of at most 2k tokens) or limited~\cite{rwkv}. The hybrid models studied here (Qwen3.5, Gemma~4, gpt-oss) as well as the recent Kimi-K3~\cite{kimik3} keep a minority of full-attention layers in an otherwise linear or windowed stack, while retaining long-context capabilities. In every case, the design decisions (ratio of full to linear layers, placement, window size of the sliding ones) are fixed by the pretraining run and baked into the released weights. These weights are the starting point of our work, not an alternative to it: we take the hybrid as given and crop only the full-attention layers that remain.

\subsection{Post-Training Linearization}

A second line keeps a released Transformer and converts its attention layers into cheaper alternatives with additional training. Hedgehog~\cite{hedgehog} learns a feature map that mimics softmax attention and converts a 7B model for a single task with LoRA; LoLCATs~\cite{lolcats} transfers attention to a feature map and recovers the accuracy with LoRA on 40M tokens for an 8B model, and scales the recipe to 70B and 405B models; Liger~\cite{liger} re-uses the key projections as gates and fine-tunes with LoRA on 20M tokens; LAWCAT~\cite{lawcat} adds a convolution to the linear attention and distills it on 94M tokens for a 7B model; SUPRA~\cite{supra} uptrains every weight on 20B tokens; and Halo~\cite{halo} distills a hybrid student from a Transformer teacher on 2.3B tokens. Jet-Nemotron~\cite{jetnemo} uses 600B tokens and 18k GPU-hours, obtaining a 12.0x speedup on Qwen2.5-1.5B. The cheapest of these conversions cost a few GPU-hours, but all of them change the weights, and only LAWCAT and Halo report long-context retrieval beyond 8k tokens (up to 22k and 256k tokens, respectively), on models of at most 8B parameters. At the frontier scales (120B, 400B) that this work targets, only LoLCATs has been applied, and its only long-context test, passkey retrieval at 8k tokens on the 8B model, fails unless the model is linearized on that same task.

\subsection{Training-Free Sparse Attention}

Like our work, these methods leave the weights untouched and change what the attention attends to. StreamingLLM~\cite{13} pins the first few tokens as attention sinks, which lets a plain sliding window keep producing coherent text past the window, but cannot retrieve information that has left it; H2O~\cite{h2o} evicts keys by accumulated attention mass; SnapKV~\cite{snapkv} selects the keys to keep at the end of the prefill. Since these selections are fixed, or only shrink, once the generation starts, they fail if the information later needed was evicted. 
Quest~\cite{quest} addresses this failure mode by re-selecting pages of the paged KV cache at every decoding step, at the cost of scoring every page at every step. Razor\-Attention~\cite{razorattention} and DuoAttention~\cite{duoattention} split the attention heads into retrieval heads, which keep the full KV cache, and streaming heads, which keep a window; Razor\-Attention identifies the retrieval heads without training, whereas DuoAttention learns one gate per head on synthetic data. Except for SnapKV (up to 47B), all of these methods were evaluated on long-context tasks with models of at most 8B parameters; DuoAttention reaches 70B, but only on short-context benchmarks.

\par
We build directly on this third line. Our contribution over it is twofold: (i) we treat the hyperparameters that these methods expose (window $W$, Quest budget $Q$, refresh interval $R$, pinned prefix size) as a \emph{per-layer} vector and search it with NSGA-II~\cite{nsga2}, rather than fixing one value globally; and (ii) we demonstrate the method on hybrids of up to 397B parameters, more than five times larger than any prior training-free report in Table~\ref{tab:related}, on long-context tasks of up to 64k tokens, and across three distinct model families.

\section{Methodology}
\label{sec:meth}

\begin{figure}
    \centering
    \includegraphics[width=\linewidth]{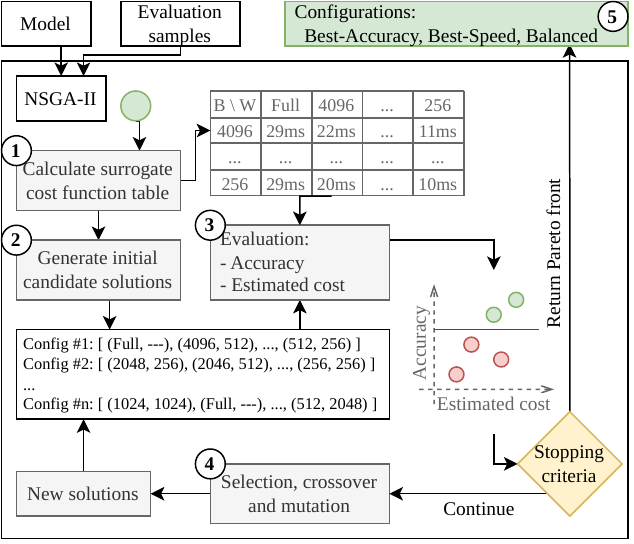}
    \vspace{-3mm}
    \caption{Overview of the pipeline used to obtain configurations using NSGA-II. (1) We fill an table with measured times with different window/budget values to be used by the surrogate cost function; (2) we generate the initial solutions; (3) the solutions are evaluated in the evaluation samples and using the cost function; (4) the best solutions are combined and mutated, and (5) when the search is over, we extract 3 configurations from the Pareto front.}
    \label{fig:pipe}
\end{figure}

Following the illustration in Figure~\ref{fig:pipe}, our pipeline receives as input a model and evaluation samples used to ensure accuracy is preserved. This method discovers new configurations using the following steps, explained in detail in this section: (1) it calculates a surrogate cost table with different window/budget pairs; (2) it generates an initial set of candidate solutions; (3) those solutions are evaluated based on their accuracy and estimated cost, and a Pareto front is calculated; (4) the best solutions are selected to be combined and mutated, generating a new set of solutions, and (5) when the run is over, we extract up to 3 solutions from the Pareto front (\textit{best-accuracy, best-speed,} and \textit{balanced}).

\subsection{Large Language Models}

\par 
As previously seen in Figure~\ref{fig:profile_context}, for long-context tasks, two kernel categories dominate the decoding time: loading weights and full attention. Since this pipeline focuses on reducing the computational cost of attention, without modifying the other aspects of decoding, users should take into consideration the loading times, as they will limit the maximum speedup obtainable by reconfiguring the LLM.

\par 
As an example, we profile the decoding of 3 models at a 64k-token context (Figure~\ref{fig:profile_decode}, in the appendix). At opposite extremes, we have Qwen3.5-4B (29.4ms in decoding, out of which 6.8 ms are loading weights) and Qwen3.5-27B (47.1ms and 19.4ms, respectively). Because this operation is unavoidable in our pipeline, the best speedup we obtained at this context length was 3.03x and 1.78x, respectively.

\par 
Given this constraint, we recommend using mixture-of-experts~(MoE) models, due to their lower number of active parameters, reducing their loading time; as well as using hybrid models, giving priority to models with higher ratios of linear- to full-attention layers, reducing our search space.

\subsection{Query-aware Sparse Attention}
\label{sec:quest}

During decoding, a full-attention layer reads the keys and values of every previous token, so the cost of each new token grows with the sequence length $L$. In the layers selected by a configuration, we replace this attention with a sparse attention that reads a bounded number of keys per step, independent of $L$, while every other layer, and every weight of the model, is left unchanged. The keys that each modified layer reads, how often they are re-selected, and what the method does and does not change are described below.

\par
\begin{figure}
    \centering
    \includegraphics[width=0.95\linewidth]{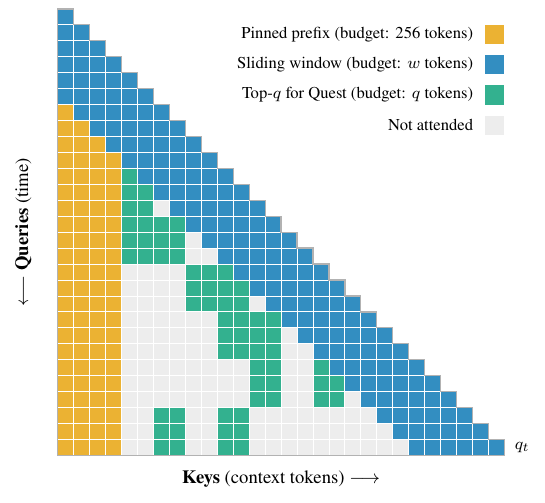}
    \vspace{-5mm}
    \caption{Combined sparse attention: one static sink (pinned prefix, 256 tokens, as $4\times\sq{sinkcol}$), sliding window ($w=6\times\sq{windowcol}$, and dynamic (Quest) sinks ($q=4 \times \sq{questcol}$), refreshed every $R$ tokens (R=3).}
    \label{fig:attention}
\end{figure}

 \textbf{From windows to query-aware pages:} A plain sliding window forgets every token that leaves the window, and preserving the first tokens as attention sinks~\cite{13} only partially recovers the accuracy lost. Information that is relevant to the current query can be anywhere in the sequence. For this reason, every modified layer attends to three sets of keys during decoding: (1) the most recent $W$ tokens (the window), (2) the first 256 tokens of the sequence, preserving part of the input~\cite{13}, and (3) the pages selected by Quest~\cite{quest}, which are the most relevant pages to the current query. The KV cache is divided into 32-token pages, which are summarized by the per-channel minimum and maximum of their keys. At each decoding step, the pages are scored, and the top pages are attended to until a budget of $Q$ tokens is used. The page selection is computed per KV head, so different heads can retrieve different parts of the sequence. By combining the three sparse-attention methods, we obtain a scheme with (1) a sliding window, (2) one static sink, and (3) dynamic sinks, updated every $R$ tokens, illustrated in Figure~\ref{fig:attention}.

\par \textbf{Refresh interval:} Scoring every page at every step is the dominant cost of Quest. We therefore recompute the selection every $R$ decoding steps, and keep it fixed in between, widening the window so that no token falls between the window and the stale selection. We use $R{=}32$ by default and refer to $R{=}1$ (selection at every step) as \emph{per-step}.

\par \textbf{Prefill and memory:} Only decoding is modified: the prefill runs vLLM's exact attention and fills the page summaries. The KV cache is kept at its full size, since any page may be selected later. As such, the method reduces the number of keys read per decoding step, and therefore the decoding time, but it does not reduce the KV cache memory footprint.

\subsection{Search for New Configurations}

Our methodology combines three sparse-attention methods and then searches for a set of parameters to minimize the model's cost while preserving its accuracy. Currently, our search algorithm is focused on minimizing two parameters: the sliding window size and the number of tokens retrieved by Quest.
Since we hypothesize that each layer has different needs, 
we use NSGA-II as a parameter search algorithm to automatically search for heterogeneous configurations. 

\subsubsection{Encoding of the Search Space}

Each full-attention layer has two components we are interested in limiting: its sliding window size $w_n$ and a Quest budget $q_n$. For a model with $N$ full-attention layers, a solution in our search space is defined as a $\big((w_1,q_1), \ldots, (w_N,q_N)\big)$ sequence, such that, $w_n \in W$ and $q_n \in Q$, where $Q$ is the possible values for Quest budget values chosen to be $\{2^8, 2^9, \ldots, 2^{12} \}$, and $W=Q \cup \{\text{full}\}$ is the set of possible sliding window sizes ($1\le n \le N$).
For example, a layer using $(w_n, q_n) = (512,256)$ will attend to the last 512 tokens, the 256 pinned tokens, and the 256 tokens selected by Quest.

\subsubsection{Obtaining configurations}
Our pipeline~(Figure~\ref{fig:pipe}) is build around using NSGA-II~\cite{nsga2} as a multi-objective search algorithm, using two objectives: maximizing accuracy and minimizing estimated cost. NSGA-II follows these steps: during initialization, NSGA-II generates a set of uniform solutions, as well as randomly generated solutions within our search space. These solutions are evaluated using the objective functions, and then solutions are selected via tournament for crossover and mutation, generating a new set of solutions, continuing the cycle. \textit{Tournament:} Two solutions are randomly selected and the dominating one (better in both objective) wins. If there is no dominating solution, the one furthest away from the other solutions in the objective space is selected (crowding distance).
\textit{Crossover:} Two solutions are split in the same index, swapping the later halves, generating 2 offspring.
\textit{Mutation:} Each parameter has a $1/n\_parameters$ chance of being multiplied or divided by 2.

\par When the search ends, we pick 3 candidate from the Pareto front (list of non-dominated solutions) that are within 3 percentage points (3pp) of the baseline model, ensuring the quality of the configurations chosen. We name these configurations \emph{Best Accuracy} (the most accurate point), \emph{Best Speed} (the point with the lowest estimated cost), and \emph{Balanced} (the point closest to the utopia point, with both objectives rescaled to $[0,1]$ over the front). 

  \begin{table}[t]
  \centering
  \caption{Total number of GPU-hours spent on the throughput (Thro.), NSGA-II, and accuracy runs.}
      \setlength{\tabcolsep}{5pt}
      \renewcommand{\arraystretch}{0.95}
      \begin{tabular}{l c c c}
      \toprule
      \textbf{Model} & \textbf{Thro.} & \textbf{NSGA-II} & \textbf{Accuracy}  \\
      \midrule
      Qwen3.5-4B            & 10.5  & 8.6   & 97.2  \\
      Qwen3.5-27B           & 22.9  & 34.2  & 465.4 \\
      Gemma-4-31B           & 38.4  & 27.5  & 693.4 \\
      gpt-oss-120b          & 13.5  & 16.2  & 220.2 \\
      \midrule
      \textbf{Total (A100)}  & 85.3  & 86.5  & 1476.2 \\
      \midrule
      Qwen3.5-122B-A10B     & 52.3  & 31.3  & 116.2  \\
      Ornith-1.0-397B       & 93.1  & 58.4 & 163.6 \\
      \midrule
      \textbf{Total (GB200)} & 145.3 & 89.7  & 279.8   \\
      \bottomrule
      \end{tabular}
  \label{tab:cost}
  \end{table}

\par \textit{Surrogate Cost Function:}
Table~\ref{tab:cost} shows the GPU-hours spent in this work, with the first columns being the time spent measuring the token/s throughput of the final solutions (up to 6 solutions per model). From these values, we conclude that measuring throughput in long-context tasks would be a very time-consuming task. We reduce the costs of measuring throughput of candidate solutions using the following expression as a low-cost surrogate cost function:
\[
\text{Cost} = T_{\text{stock}} - \sum_{n=1}^{N} \big( T_{\text{full}} - T_{\text{quest}}(w_n, q_n) \big),
\]
where $T_{\text{stock}}$ is the measured time of the base model, $T_{\text{full}}$ is the time spent in one full-attention layer, and $T_{\text{quest}}(w, q)$ is the time spent in a layer with window $w$ and budget $q$ ($T_{\text{quest}} = T_{\text{full}}$ for an unmodified layer). The values of $T_{\text{quest}}$ are measured for every $(w, q)$ pair by measuring the time it takes to go through a layer with that configuration 64k times using batch size 8.

  \begin{table*}[t]
      \centering
      \caption{Models studied in this work and hardware used. For multi-GPU runs, we use tensor parallelism of 2 and 4.}
      \setlength{\tabcolsep}{3pt}
      \begin{tabular}{l c c c c c c c c}
      \toprule
       & & \multicolumn{2}{c}{Parameters} & & & & & \\
      \cmidrule(lr){3-4}
      Model & Type & Total & Active & Layers & Full attn. & Other layers & Context & Hardware \\
      \midrule
      Qwen3.5-4B            & dense & 4B    & 4B    & 32 & 8  & GDN        & 262k & 1$\times$A100 \\
      Qwen3.5-27B           & dense & 27B   & 27B   & 64 & 16 & GDN        & 262k & 1$\times$A100 \\
      Qwen3.5-122B-A10B-FP8 & MoE   & 122B  & 10B   & 48 & 12 & GDN        & 262k & 4$\times$GB200 \\
      Ornith-1.0-397B-FP8   & MoE   & 397B  & 17B   & 60 & 15 & GDN        & 262k & 4$\times$GB200 \\
      Gemma-4-31B           & dense & 30.7B & 30.7B & 60 & 10 & SWA-1024   & 262k & 2$\times$A100 \\
      gpt-oss-120b          & MoE   & 117B  & 5.1B  & 36 & 18 & SWA-128    & 131k & 2$\times$A100 \\
      \bottomrule
      \end{tabular}
      \label{tab:models}
  \end{table*}

\section{Evaluation}
\label{sec:results}

While our pipeline uses a search algorithm to automatically find configurations, we acknowledge that it is possible to search for configurations manually, despite that placing a cost in deployment that is measurable in human-hours. As such, we compare two ways of obtaining a configuration: a uniform configuration with parameters decided through manual trial and error, and the configurations returned by our search algorithm.

\par \textbf{Uniform configuration (manual):} Every full-attention layer uses the same window $W{=}256$ and budget $Q{=}1024$, with the 256 pinned tokens and a refresh interval of $R{=}32$ (R32), or $R{=}1$ (R1, per-step). These values are the result of a long sequence of manual experiments, in which we first tried plain sliding windows, then windows with sinks, SnapKV-placed sinks, and Quest with different page sizes, budgets, and refresh intervals. Its main limitation is that it applies the same configuration to every layer and every model, despite our assumption that different models and layers have different needs.

\par \textbf{NSGA-II (automatic):} We use NSGA-II~\cite{nsga2} as a multi-objective search algorithm, using two objectives: maximizing accuracy and minimizing the estimated cost. The accuracy objective is the mean score on two long-context datasets: BABILong-qa1 at 64k tokens~\cite{babilong} and LongBench Qasper~\cite{longbench}. Each candidate is evaluated on the first 100 examples of each dataset. The initial population is seeded with three uniform configurations ($W{=}256$; $Q \in \{512, 1024, 2048\}$), and the rest is sampled at random. The runs were targeted to fit into a 15-hour wall time block, resulting in the following number of iterations (it.) and population size (ps) per setup: 1xA100 (4 it., ps 10), 2xA100 (3 it., ps 8), 4xB200 (3 it., ps 10).

\par \textbf{Hybrid models:} We apply our methodology to six representative hybrid models from three families (Table~\ref{tab:models}). In every model, only a minority of the decoder layers compute a full softmax-attention matrix; the remaining layers are either linear attention (Gated DeltaNet~(GDN)~\cite{7} in the Qwen3.5 family~\cite{11,12}) or sliding-window attention (Gemma~4~\cite{gemma4} and gpt-oss~\cite{gptoss}). Since those layers are already $O(L)$ or bounded by their own window, they are not modified in any way in this study: the search space is exactly the set of full-attention layers.

\par 
In the remainder of this section, we discuss the configurations discovered through \mbox{NSGA-II}, followed by the speedups obtained by applying both the uniform and \mbox{NSGA-II} configurations, and finally the accuracy on the long-context benchmark, showing that this methodology is applied without affecting the models' accuracy.

\subsection{Configurations Discovered Through NSGA-II}
\label{sec:phaseB}

\par
The NSGA-II algorithm is used to search for configurations that maximize accuracy while minimizing their estimated decoding cost. 
The configurations obtained have been placed in the appendix (Section~\ref{app:configs}). After running the search algorithm, we obtain up to three configurations, all within 3pp of the baseline accuracy: \textit{Best-Accuracy}, \textit{Best-Speed}, and \textit{Balanced}, which are described below.

\par
\textbf{Qwen3.5-4B:} \textit{Best-Accuracy} keeps only one layer at full attention, and crops the other 7 layers with windows between 256 and 4096 and budgets between 256 and 2048. Balanced crops all 8 layers, with a window of 256 on 6 of them, and Best Speed crops all 8 layers to $W{=}256$, $Q{=}512$.

\par
\textbf{Qwen3.5-27B:} Every pick crops all 16 layers. Best Speed uses $W{=}256$, $Q{=}512$ on every layer, and Balanced differs from it only in a budget of 2048 on layer 7. Best Accuracy keeps a window of 256 on 11 layers, widens it to 2048--4096 on the other 5 (layers 3, 31, 51, 55 and 59), and uses budgets between 256 and 4096.

\par
\textbf{Gemma-4-31B:} \textit{Best-Accuracy} crops only 2 of the 10 layers (layer 11 to $W{=}1024$ and layer 59 to $W{=}4096$), and Best Speed crops 6, keeping layers 5, 35, 41 and 47 at full attention. The \textit{Balanced} point of the front coincides with \textit{Best-Accuracy}, so it is not reported separately.

\par
\textbf{gpt-oss-120b:} The front contained a single configuration, which crops 4 of the 18 layers (layers 3, 5, 25 and 33), and is therefore reported as \textit{Best-Accuracy} only.

\par
\textbf{Qwen3.5-122B:} The front contained 4 configurations, and the three picks are close to each other: on the 21 datasets, their average accuracy is within 0.2~pp of each other, and their speed is within 1\%.

\par
\textbf{Ornith-397B:} Best Speed is a uniform configuration ($W{=}256$, $Q{=}1024$ on all 15 layers), while Best Accuracy and Balanced use different windows and budgets per layer. The front contained 3 configurations, and the uniform configuration had a budget of 256.

Considering that out top-3 models have 8 to 15 full-attention layers, there are $25^8$--$25^{15}$ possible configurations, and we only try 24--40 configurations per model. We leave a large portion of the search space unexplored, which LLMs could explore more efficiently as decision makers~\cite{Iba2026}. Despite the limited number of explored solutions, it is interesting to say that the majority of the configurations cropped all or most full-attention layers, indicating the current models can still be vastly improved.

\subsection{Controlled throughput benchmark}
\label{sec:throughput-results}
\label{sec:throughput}

The wall-clock time from the accuracy runs cannot tell us how much faster the model has become. In those runs, the output length is a free variable, making the measured time dependent on the length of the answer, and not on the speed of the model. As such, we make a controlled throughput benchmark, in which we measure how long each configuration (\textit{baseline, best-accuracy, best-speed, balanced, R32} and \textit{R1}) performs the same work at different batch sizes. 

\par 
For each batch size $B \in \{1, 4, 16\}$, we generated $B$ synthetic 128-token prompts, and every request generates exactly $L_{\max}$ tokens (end-of-sequence ignored). The engine is stepped manually, and the wall-clock time $t(L)$ at which every request in the batch has produced $L$ tokens is recorded for $L = 4\text{k}, 8\text{k}, 16\text{k}, \dots, 256\text{k}$. The throughput at length $L$ is defined as $(B \cdot L) / t(L)$. A batch size stops at the largest checkpoint whose KV cache fits in memory ($B \cdot (128 + L)$ within 90\% of the capacity vLLM reports. The results are presented as one figure per model with the top-3 best speedups in Figure~\ref{fig:throughput_main}, and all models in the appendix (Figure~\ref{fig:throughput}). Table~\ref{tab:throughput-summary} shows the values for batch size 4.

\begin{figure}[!]
  \centering
  Single-GPU setup: A100\\
\includegraphics[width=0.94\linewidth,trim=0 25 0 0,clip]{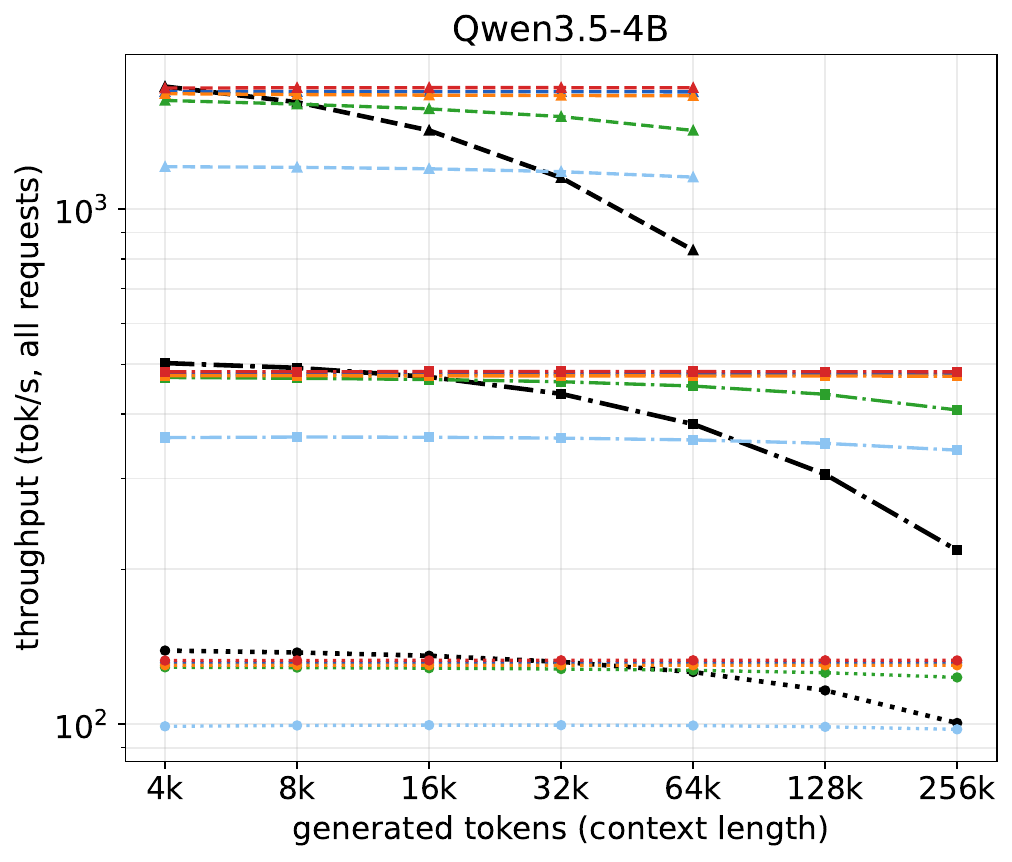}
  Quad-GPU setup: GB200 node\\
\includegraphics[width=0.94\linewidth,trim=0 25 0 0,clip]{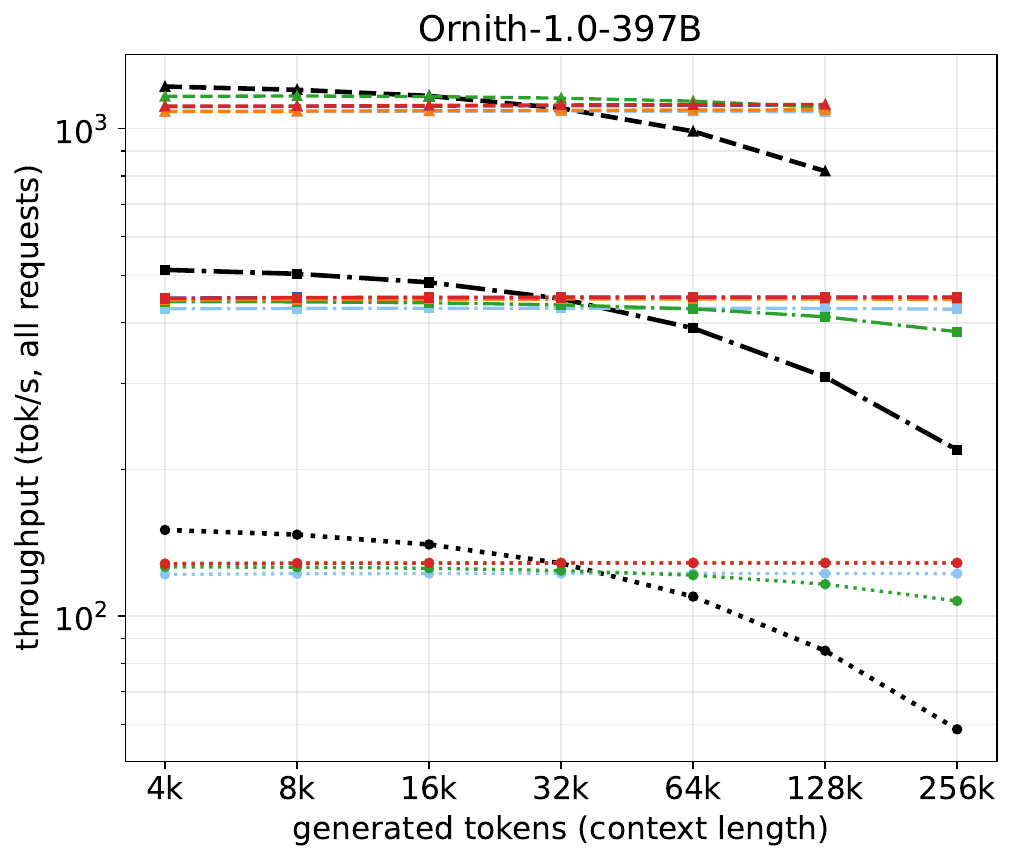}\\
\includegraphics[width=0.94\linewidth]{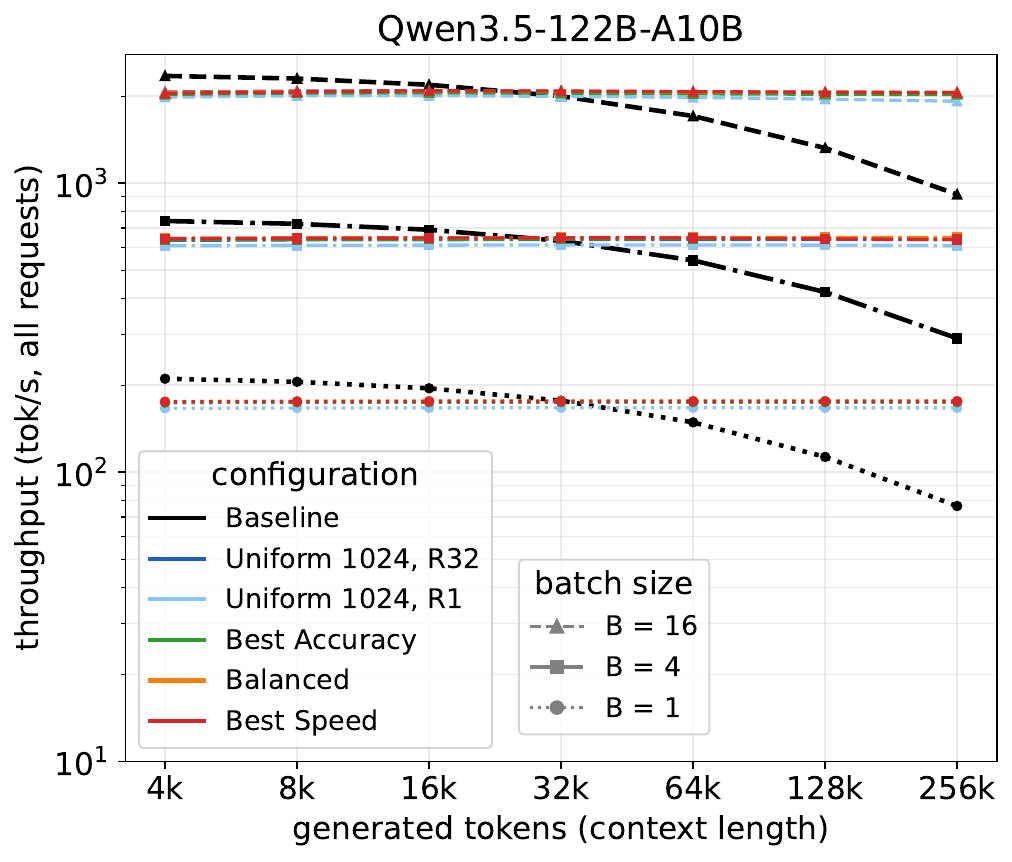}\\
  \caption{Token/s throughput using different configurations and batch sizes.
 The baseline models are faster for short sequences due to the computation overhead from Quest. For longer sequences starting from 64k tokens, we see reasonable speedups in Qwen-4B, Qwen-122B, and Ornith. These speedups range 1.16--1.26x in context lengths of 64k tokens and 2.06--2.23x for 256k tokens}
  \label{fig:throughput_main}
  \end{figure}
\begin{table}[!t]
    \centering
    \caption{Decode speedup over the baseline at the same context length and batch size 4, followed by the throughput in brackets (tokens/s, all requests). For each model, we show the results of the baseline, \textit{Best Accuracy}, the fastest between \textit{Balance} and \textit{Best Speed} configurations, and R32 (applying a uniform 1024-token budget to all layers, updating quest every 32 steps). The GPT-OSS run returned a single configuration. Ornith's Best Speed and R32 configurations are the same.
    `---' marks a point that was not measured either because the context is beyond the model's limit (GPT-OSS) or the KV cache would not fit into memory at batch size 4 (Qwen-27B and Gemma runs).
    }
    \footnotesize
    \setlength{\tabcolsep}{3pt}
    \begin{tabular}{l ccc}
    \toprule
    \textbf{Model /} & \multicolumn{3}{c}{\textbf{Generated tokens}} \\
    \cmidrule(lr){2-4}
    \textbf{Configuration} & 16k & 64k & 256k \\
    \midrule
    \multicolumn{4}{l}{\textbf{Qwen3.5-4B}} \\
    \hspace{0.5em}Baseline & \hspace{2.5em} 472.3 & \hspace{2.5em} 382.4 & \hspace{2.5em} 217.6 \\
    \hspace{0.5em}Best Accuracy & 0.99$\times$ {\scriptsize(466.8)} & 1.19$\times$ {\scriptsize(453.4)} & 1.87$\times$ {\scriptsize(407.3)} \\
    \hspace{0.5em}Best Speed & 1.02$\times$ {\scriptsize(483.5)} & 1.26$\times$ {\scriptsize(483.5)} & 2.22$\times$ {\scriptsize(483.0)} \\
    \hspace{0.5em}R32 & 1.01$\times$ {\scriptsize(478.3)} & 1.25$\times$ {\scriptsize(478.2)} & 2.20$\times$ {\scriptsize(477.7)} \\
    \midrule
    \multicolumn{4}{l}{\textbf{Qwen3.5-27B}} \\
    \hspace{0.5em}Baseline & \hspace{2.5em} 106.8 & \hspace{2.5em} 96.7 & --- \\
    \hspace{0.5em}Best Accuracy & 0.97$\times$ {\scriptsize(103.4)} & 1.07$\times$ {\scriptsize(103.4)} & --- \\
    \hspace{0.5em}Balanced & 1.00$\times$ {\scriptsize(107.0)} & 1.11$\times$ {\scriptsize(107.0)} & --- \\
    \hspace{0.5em}R32 & 0.97$\times$ {\scriptsize(104.0)} & 1.08$\times$ {\scriptsize(104.0)} & --- \\
    \midrule
    \multicolumn{4}{l}{\textbf{Gemma-4-31B}} \\
    \hspace{0.5em}Baseline & \hspace{2.5em} 141.0 & \hspace{2.5em} 122.3 & --- \\
    \hspace{0.5em}Best Accuracy & 0.99$\times$ {\scriptsize(139.5)} & 1.01$\times$ {\scriptsize(123.3)} & --- \\
    \hspace{0.5em}Best Speed & 0.99$\times$ {\scriptsize(139.4)} & 1.05$\times$ {\scriptsize(127.9)} & --- \\
    \hspace{0.5em}R32 & 0.99$\times$ {\scriptsize(139.5)} & 1.09$\times$ {\scriptsize(133.1)} & --- \\
    \midrule
    \multicolumn{4}{l}{\textbf{gpt-oss-120b}} \\
    \hspace{0.5em}Baseline & \hspace{2.5em} 529.8 & \hspace{2.5em} 382.3 & --- \\
    \hspace{0.5em}Best Acc. / Speed & 0.87$\times$ {\scriptsize(461.3)} & 0.97$\times$ {\scriptsize(371.6)} & --- \\
    \hspace{0.5em}R32 & 0.61$\times$ {\scriptsize(322.9)} & 0.84$\times$ {\scriptsize(322.0)} & --- \\
    \midrule
    \multicolumn{4}{l}{\textbf{Qwen3.5-122B-A10B-FP8}} \\
    \hspace{0.5em}Baseline & \hspace{2.5em} 690.2 & \hspace{2.5em} 540.8 & \hspace{2.5em} 290.8 \\
    \hspace{0.5em}Best Accuracy & 0.93$\times$ {\scriptsize(640.0)} & 1.19$\times$ {\scriptsize(642.4)} & 2.21$\times$ {\scriptsize(642.9)} \\
    \hspace{0.5em}Balanced & 0.94$\times$ {\scriptsize(647.1)} & 1.20$\times$ {\scriptsize(648.2)} & 2.23$\times$ {\scriptsize(648.3)} \\
    \hspace{0.5em}R32 & 0.92$\times$ {\scriptsize(637.9)} & 1.18$\times$ {\scriptsize(639.0)} & 2.20$\times$ {\scriptsize(639.3)} \\
    \midrule
    \multicolumn{4}{l}{\textbf{Ornith-1.0-397B-FP8}} \\
    \hspace{0.5em}Baseline & \hspace{2.5em} 484.0 & \hspace{2.5em} 389.8 & \hspace{2.5em} 219.4 \\
    \hspace{0.5em}Best Accuracy & 0.91$\times$ {\scriptsize(439.3)} & 1.10$\times$ {\scriptsize(426.9)} & 1.75$\times$ {\scriptsize(383.3)} \\
    \hspace{0.5em}Best Speed / R32 & 0.93$\times$ {\scriptsize(450.5)} & 1.16$\times$ {\scriptsize(451.0)} & 2.06$\times$ {\scriptsize(450.9)} \\
    \bottomrule
    \end{tabular}
    \label{tab:throughput-summary}
\end{table}

\par \textbf{Short contexts} are slower. While Quest allows us to save on computation by only considering the top-$k$ tokens, this also introduces an additional computational overhead that makes the configurations slower than the baseline for context lengths up to 16k--32k tokens (excluding gpt-oss-120b). This motivated us to only rerank the Quest pages once every 32 tokens, vastly reducing the computational overhead that results in the \textit{R1} configuration being the slowest.

\par \textbf{Long contexts} are faster. In full-attention blocks, the cost of generating each token grows linearly with the context length, making each token more computationally expensive than the previous one. Since our approach has a computational cost independent of the context length (despite KV cache still growing normally), Quest's computation overhead is overshadowed by the full-attention costs in long contexts. At the longest context, the speedups reach 1.26/2.22$\times$ on Qwen3.5-4B, 1.20/2.23$\times$ on Qwen3.5-122B-A10B, and 1.16/2.06$\times$ on Ornith-1.0-397B, for batch size 4.

\par \textbf{Computational overhead} of Quest. Quests need to rank all attention pages every decode step and every layer, or every 32 decode steps if you follow our methodology. This adds an overhead that is highly dependent on the number of full-attention layers within the model. When we profiled our configuration during decode (Figure~\ref{fig:profile_decode}, in appendix), we noticed that this overhead is minimal, especially when done only once every 32 decode steps.

\par \textbf{Baseline model bottleneck:} This method was not very effective in the Qwen3.5-27B and Gemma-4-31B models. Both of these methods are dense, resulting in a reasonable amount of time being the time spent reading weights, making the savings on compute less noticeable in the throughput. Similarly, Qwen3.5-4B is also a dense model, but due to its small size, the time spent reading parameters is small enough to obtain a good speedup from optimizing the computational side of decode.

Across the 3 models with the best speedups, we obtained speedups in the range of 1.16--1.26x in context lengths of 64k tokens and 2.06--2.23x when pushing the context length to 256k tokens. In addition to this, it should be noted that the configuration in which Quest is applied to all layers became constant in terms of token/s produced. As will be show next, these configurations did not degrade the accuracy of the baseline models, making these speedups more impressive.

\subsection{Accuracy in Long-context tasks}
\label{sec:accuracy}

We evaluate the configurations on 21 long-context datasets: (1) BABILong qa1--qa5~\cite{babilong} at 64k tokens; and (2) the 16 English subsets of LongBench~\cite{longbench} (single- and multi-document QA, summarisation, few-shot learning, synthetic retrieval and code completion). We use the full test set of every dataset, \textbf{excluding} the 100 examples of BABILong-qa1 and Qasper that were used during the search, so the reported accuracy is measured on held-out data.

\par \textbf{Protocol:} Each configuration is compared with the baseline model. Decoding is greedy, with thinking disabled and the maximum number of new tokens defined per dataset (256 for BABILong, and LongBench's own limits). LongBench is scored with its official metrics (F1, ROUGE-L, accuracy, and edit similarity for code), and BABILong by answer match. Before scoring, typographic characters (narrow and thin spaces or commas between digit groups, non-breaking hyphens and curly quotes) are normalized to ASCII, as some models emit them in place of the characters in the references. A configuration is considered to preserve accuracy on a dataset if its score is within 3pp of the baseline.

\par \textbf{Prompts:} Each model receives its own chat template. For gpt-oss, which uses the harmony format, the assistant turn is opened directly in its \emph{final} channel, so the answer budget is not spent on its analysis channel, and LongBench's few-shot and code-completion subsets, which LongBench runs as raw completions, are also wrapped in the chat template.

\begin{sidewaystable}[p]
\centering
\caption{
Accuracy (\%) in each dataset and their average. Base = baseline, Acc / Bal / SpdB = accuracy-best, balanced, and speed-best configurations, R32 = uniform window 256 with Quest budget 1024 re-selected every 32 decode steps, R1 the same at every step. \colorbox{cellRed}{Red}/\colorbox{cellGreen}{green} mark a drop and a gain of more than 1pp against the baseline.}
\setlength{\tabcolsep}{2pt}
\renewcommand{\arraystretch}{0.8}
\resizebox{0.92\textheight}{!}{%
\begin{tabular}{l |cccccc|cccccc|cccc|ccccc|cccccc|cccc}
\toprule
 & \multicolumn{6}{c}{\textbf{Qwen-4B}} & \multicolumn{6}{c}{\textbf{Qwen-27B}} & \multicolumn{4}{c}{\textbf{gpt-oss}} & \multicolumn{5}{c}{\textbf{Gemma-31B}} & \multicolumn{6}{c}{\textbf{Qwen-122B}} & \multicolumn{4}{c}{\textbf{Ornith-397B}} \\
\cmidrule(lr){2-7}\cmidrule(lr){8-13}\cmidrule(lr){14-17}\cmidrule(lr){18-22}\cmidrule(lr){23-28}\cmidrule(lr){29-32}
\textbf{Dataset} & Base & Acc & Bal & Spd & R32 & R1 & Base & Acc & Bal & Spd & R32 & R1 & Base & Acc & R32 & R1 & Base & Acc & Spd & R32 & R1 & Base & Acc & Bal & Spd & R32 & R1 & Base & Acc & Bal & Spd/R32 \\
\midrule
qa1 & 61.0 & 60.9 & 60.8 & 60.7 & 60.9 & 60.7 & 92.2 & 91.9 & 92.1 & 92.1 & 92.0 & 92.1 & 47.6 & 47.3 & 47.3 & 47.2 & 75.0 & 74.8 & 74.8 & 74.8 & 74.8 & 87.2 & 87.2 & 87.2 & 87.1 & 87.1 & 87.2 & 85.6 & 85.6 & 85.6 & 85.6 \\
qa2 & 38.4 & 38.3 & 38.2 & 38.1 & 38.2 & 38.4 & 50.7 & 51.2 & 50.4 & 50.5 & 49.8 & 50.7 & 29.0 & 29.2 & \colorbox{cellGreen}{30.2} & 29.8 & 41.4 & 41.7 & 41.8 & 41.5 & 41.5 & 44.0 & 43.9 & 43.9 & 43.9 & 43.9 & 43.9 & 32.9 & 31.9 & \colorbox{cellRed}{31.3} & \colorbox{cellRed}{31.7} \\
qa3 & 26.3 & 26.1 & 26.2 & 26.2 & 25.9 & 26.1 & 30.3 & 30.5 & 30.4 & 30.7 & 30.5 & 30.7 & 26.0 & 26.0 & 25.8 & 25.6 & 28.2 & 27.4 & 27.4 & 27.4 & 27.4 & 32.3 & 32.3 & 32.3 & 32.3 & 32.3 & 32.3 & 8.4 & \colorbox{cellRed}{7.1} & 7.8 & \colorbox{cellRed}{5.5} \\
qa4 & 59.7 & 59.5 & 59.6 & 59.5 & 59.5 & 59.5 & 57.1 & 57.0 & 56.7 & 56.9 & 57.1 & 56.9 & 45.4 & 45.0 & 44.4 & 45.5 & 85.9 & 86.4 & 86.4 & 86.4 & 86.4 & 67.5 & 67.5 & 67.5 & 67.5 & 67.5 & 67.5 & 61.3 & 61.3 & 61.3 & 61.3 \\
qa5 & 72.1 & 72.0 & 72.0 & 72.0 & 72.0 & 72.0 & 80.4 & 80.5 & 80.6 & 80.6 & 80.3 & 80.4 & 61.7 & 62.0 & 62.0 & 61.8 & 81.5 & 81.6 & 81.6 & 81.6 & 81.6 & 83.9 & 83.9 & 83.9 & 83.9 & 83.9 & 83.9 & 84.7 & 84.7 & 84.7 & 84.7 \\
Mus & 44.9 & 45.3 & 44.8 & 44.8 & 45.4 & 45.4 & 55.3 & 54.6 & 54.8 & 55.7 & 55.5 & 55.5 & 33.6 & 33.3 & \colorbox{cellRed}{32.5} & 33.2 & 52.7 & 52.5 & 52.5 & 52.5 & 52.5 & 57.2 & 57.2 & 57.2 & 57.2 & 57.2 & 57.2 & 57.1 & 57.1 & 57.1 & 57.2 \\
HQA & 64.7 & 64.6 & 64.8 & 64.4 & 64.6 & 65.1 & 67.4 & 67.7 & 67.6 & 67.5 & 67.4 & 68.1 & 49.0 & \colorbox{cellGreen}{51.5} & \colorbox{cellGreen}{51.7} & \colorbox{cellGreen}{50.5} & 67.3 & 68.0 & 68.0 & 68.0 & 68.0 & 70.6 & 70.6 & 70.5 & 70.5 & 70.5 & 70.5 & 69.3 & 69.3 & 69.4 & 69.4 \\
2Wiki & 57.4 & 57.4 & 57.3 & 57.8 & 57.4 & 57.6 & 67.9 & 68.0 & \colorbox{cellRed}{66.7} & 67.8 & 67.5 & 67.9 & 38.4 & 38.6 & 39.0 & 38.9 & 71.4 & 71.9 & 71.9 & 71.9 & 71.9 & 70.5 & 70.5 & 70.5 & 70.5 & 70.5 & 70.5 & 73.2 & 73.2 & 73.2 & 73.0 \\
NQA & 30.1 & 30.0 & 30.0 & 30.1 & 29.9 & 29.3 & 33.3 & 33.2 & 33.3 & 33.2 & 33.2 & 33.8 & 23.2 & \colorbox{cellGreen}{24.8} & \colorbox{cellGreen}{24.6} & 23.2 & 32.5 & 32.5 & 32.5 & 32.5 & 32.6 & 32.6 & 32.6 & 32.6 & 32.6 & 32.6 & 32.6 & 33.2 & 33.2 & 33.2 & 33.2 \\
Qas & 46.3 & 45.7 & \colorbox{cellRed}{44.0} & \colorbox{cellRed}{43.9} & 45.8 & 45.9 & 50.7 & 49.9 & \colorbox{cellRed}{48.1} & \colorbox{cellRed}{49.2} & 50.5 & 51.2 & 37.5 & 37.6 & \colorbox{cellRed}{36.0} & 37.4 & 51.4 & 51.7 & 51.6 & \colorbox{cellRed}{48.4} & \colorbox{cellRed}{49.4} & 47.2 & \colorbox{cellGreen}{48.6} & \colorbox{cellGreen}{48.6} & 47.0 & \colorbox{cellGreen}{48.8} & 47.9 & 49.7 & 49.9 & \colorbox{cellRed}{47.8} & 49.9 \\
GovR & 31.9 & 31.0 & \colorbox{cellRed}{29.4} & \colorbox{cellRed}{27.9} & \colorbox{cellRed}{30.4} & 31.8 & 32.1 & \colorbox{cellRed}{30.9} & \colorbox{cellRed}{29.2} & \colorbox{cellRed}{29.2} & \colorbox{cellRed}{30.7} & 31.7 & 22.2 & 22.1 & 21.4 & 21.9 & 28.3 & 28.4 & 28.1 & 27.4 & 27.8 & 30.8 & 30.2 & 30.2 & 30.6 & 30.2 & 31.0 & 29.9 & 30.0 & 29.3 & 29.1 \\
LCC & 52.7 & 52.0 & \colorbox{cellRed}{51.4} & \colorbox{cellRed}{51.4} & 52.3 & 51.8 & 67.5 & \colorbox{cellGreen}{68.6} & \colorbox{cellGreen}{68.6} & 67.2 & 67.4 & 67.2 & 10.4 & 10.5 & 10.2 & 10.1 & 6.0 & 6.0 & 5.5 & 6.1 & 6.2 & 68.1 & 68.8 & 67.7 & 68.5 & 68.9 & 68.6 & 76.1 & 76.0 & 75.9 & 76.2 \\
MNews & 23.8 & 23.8 & 23.6 & 22.8 & 23.9 & 23.8 & 23.7 & 22.8 & 23.2 & 22.9 & 23.4 & 23.3 & 21.4 & 21.5 & 21.1 & 21.5 & 23.9 & 23.8 & 23.7 & 23.6 & 23.8 & 23.9 & 23.5 & 23.6 & 23.6 & 23.8 & 23.8 & 24.4 & 24.4 & 23.9 & 23.9 \\
MFQA & 57.9 & 58.3 & 57.6 & \colorbox{cellRed}{56.8} & 58.0 & 58.7 & 55.2 & 56.0 & 54.2 & \colorbox{cellRed}{54.1} & 55.2 & 55.4 & 51.2 & \colorbox{cellRed}{49.8} & \colorbox{cellRed}{50.0} & \colorbox{cellRed}{49.0} & 57.7 & 57.6 & 57.9 & 57.6 & 57.2 & 55.9 & 56.0 & 55.5 & \colorbox{cellGreen}{57.2} & 55.7 & 56.3 & 56.0 & \colorbox{cellGreen}{57.1} & 56.5 & 55.5 \\
PCnt & 8.5 & 8.5 & 8.5 & 8.5 & 8.5 & 8.5 & 16.0 & 16.0 & 17.0 & 16.5 & 15.5 & 16.5 & 8.5 & 9.0 & 8.0 & \colorbox{cellRed}{7.0} & 0.5 & 0.9 & 0.7 & 0.9 & 0.9 & 21.0 & 21.0 & 21.0 & 21.0 & 21.0 & 21.0 & 13.7 & 13.7 & 13.7 & 13.7 \\
PRet & 92.5 & 92.5 & \colorbox{cellRed}{91.3} & 91.8 & 92.6 & 92.4 & 100.0 & 100.0 & 100.0 & 100.0 & 100.0 & 100.0 & 99.0 & 100.0 & 99.5 & 99.0 & 100.0 & 100.0 & 100.0 & 100.0 & 100.0 & 100.0 & 100.0 & 100.0 & 100.0 & 100.0 & 100.0 & 100.0 & 100.0 & 100.0 & 100.0 \\
QMS & 22.0 & 21.7 & 21.4 & 21.1 & 21.4 & 22.5 & 22.2 & 21.8 & 21.4 & 21.4 & 21.9 & 21.9 & 19.6 & 20.0 & 19.1 & 19.4 & 24.2 & 24.1 & 24.1 & 23.5 & 23.8 & 22.6 & 22.1 & 22.0 & 22.1 & 22.1 & 22.4 & 22.4 & 23.0 & 22.1 & 21.9 \\
Repo & 49.1 & 48.6 & \colorbox{cellRed}{47.6} & \colorbox{cellRed}{45.9} & \colorbox{cellRed}{47.9} & 48.3 & 51.6 & \colorbox{cellGreen}{54.5} & \colorbox{cellGreen}{54.2} & \colorbox{cellGreen}{55.4} & 51.8 & \colorbox{cellRed}{49.6} & 13.4 & 13.7 & \colorbox{cellRed}{12.3} & 12.8 & 15.6 & 16.0 & 15.6 & \colorbox{cellRed}{13.7} & \colorbox{cellRed}{13.8} & 66.1 & \colorbox{cellRed}{65.0} & 65.7 & 66.0 & \colorbox{cellRed}{64.5} & \colorbox{cellRed}{65.0} & 71.9 & 72.7 & \colorbox{cellRed}{70.6} & 71.0 \\
SamS & 34.6 & 35.2 & 35.3 & 34.4 & 35.1 & 35.3 & 40.5 & \colorbox{cellRed}{39.0} & \colorbox{cellRed}{39.3} & \colorbox{cellRed}{38.8} & 39.9 & 39.9 & 26.4 & 27.4 & \colorbox{cellGreen}{27.6} & 26.8 & 23.5 & 24.4 & 22.9 & 22.6 & 22.8 & 37.5 & \colorbox{cellRed}{36.0} & 37.3 & 36.9 & 37.1 & \colorbox{cellRed}{35.7} & 13.3 & \colorbox{cellRed}{11.8} & \colorbox{cellRed}{12.0} & 12.3 \\
TREC & 64.0 & \colorbox{cellRed}{61.0} & 64.0 & \colorbox{cellRed}{56.0} & \colorbox{cellRed}{61.0} & \colorbox{cellRed}{58.5} & 49.5 & \colorbox{cellGreen}{55.5} & \colorbox{cellGreen}{51.0} & \colorbox{cellGreen}{58.0} & \colorbox{cellGreen}{54.5} & \colorbox{cellGreen}{55.0} & 51.5 & 51.5 & 52.5 & \colorbox{cellGreen}{53.0} & 63.5 & 62.5 & 63.0 & 64.5 & 63.5 & 75.0 & 74.5 & 74.0 & 74.0 & 74.0 & 74.5 & 82.5 & 82.0 & 82.0 & 82.0\\
TQA & 87.1 & 87.1 & 86.2 & 86.8 & 86.6 & 87.2 & 86.5 & \colorbox{cellRed}{84.4} & \colorbox{cellRed}{83.1} & \colorbox{cellRed}{84.2} & \colorbox{cellRed}{85.4} & \colorbox{cellRed}{84.9} & 82.3 & 81.4 & \colorbox{cellRed}{81.0} & 81.9 & 83.7 & 84.0 & 83.8 & 82.7 & 83.1 & 92.1 & 92.6 & 92.1 & 92.2 & 91.6 & 92.5 & 88.2 & 89.1 & \colorbox{cellGreen}{89.4} & 88.2 \\
\midrule
\textbf{Avg} & \textbf{48.8} & \textbf{48.5} & \textbf{48.3} & \colorbox{cellRed}{\textbf{47.7}} & \textbf{48.4} & \textbf{48.5} & \textbf{53.8} & \textbf{54.0} & \textbf{53.4} & \textbf{53.9} & \textbf{53.8} & \textbf{53.9} & \textbf{38.0} & \textbf{38.2} & \textbf{37.9} & \textbf{37.9} & \textbf{48.3} & \textbf{48.4} & \textbf{48.3} & \textbf{48.0} & \textbf{48.0} & \textbf{56.5} & \textbf{56.4} & \textbf{56.3} & \textbf{56.4} & \textbf{56.3} & \textbf{56.4} & \textbf{54.0} & \textbf{54.0} & \textbf{53.7} & \textbf{53.6} \\
\bottomrule
\end{tabular}}
\label{tab:phaseDflip1}
\end{sidewaystable}


Table~\ref {tab:phaseDflip1} shows the accuracy values for each configuration on the 21 long-context datasets. In this table, we highlight in \green{green} and \red{red} values that are 1 percentage point above or below the baseline, and consider a configuration successful if they do not degrade the results of the baseline model on any of the datasets by more than 3pp. The later comments will be focused on Qwen-4B, Qwen-122B, and Ornith, since they obtained the biggest speedup.

\par \textbf{No visible loss in accuracy:} These values show that our experiments were successful. The \textit{Best Accuracy} and \textit{R32} configurations showed no accuracy degradation in any of the models and datasets. The \textit{Best Speed} configuration obtained the best throughput while only reducing accuracy in three datasets when applied to the Qwen-4B model.

\par \textbf{Qwen3.5-4B:} The Best Accuracy, Balanced, and R32 configurations showed no degradation in accuracy in any individual dataset, and their average accuracy stayed within 0.5 percentage points of the baseline model. This indicates that our 1.19x--1.26x (64k tokens) and 1.87--2.22x (256k tokens) speedups do not bring down the quality of the model.

\par \textbf{Qwen3.5-122B:} All of the final configurations tested using this model passed all the 21 long-context datasets, and their average accuracy stayed within 0.2 percentage points of the baseline. Once again, this validates that this method can reliably speed up models. Similarly to the previous model, we obtained 1.18--1.20x speedups in contexts with 64k tokens and 2.20--2.23x for 256k tokens.

\par \textbf{Ornith-397B:} Similarly to the previous model, there is no accuracy degradation in individual benchmarks, and the average accuracy of the configurations stays within 0.4pp of the baseline. In addition to the configurations tested in the previous models, we try pushing the uniform configuration to a lower budget, damaging the accuracy in 4 datasets.

The results show that this combination of pinned prefix + Quest + sliding window attention is an effective method to accelerate models in long-context tasks with little to no damage to the models' capabilities.

\subsection{Testing Qwen3.5-4B for Code Optimization}

\par
This method shows good results at preserving accuracy in long-context benchmarks, whether the configurations were obtained manually or through NSGA-II. We decided to extend the evaluation to kernel optimization using the PolyBench repository~\cite{poly}. In this task, we attempt to optimize 4 kernels (\textit{gemm, correlation, heat-3d} and \textit{lu}) in parallel using batch size 4, and a sequence of requests until the context length reaches 256k tokens. The use case is detailed in the appendix~(Section~\ref{poly}).

\par 
For this optimization task, we compare the Qwen3.5-4B model with 3 configurations: \textit{Best-Accuracy} and two uniform configurations using $(w, q) = 2048$ and $4096$. We do a series of prompts to the model, evaluating the proposed optimization at the end of each prompt. Prompt 1 described the problem; prompts 2-4 recommended focusing on memory access order; prompts 5-8 loop ordering and tiling; prompts 9-12 OpenMP; prompts 13-16 SIMD and vectorization; prompts 17-20 combining previous implementations and retuning parameters, and all prompts until reaching a 256k-token context alternated between asking the model for a different loop structure and tuning parameters.

\par 
For these kernels, the baseline model sped up 3 kernels (\textit{gemm}, \ textit {correlation}, and \textit{heat-3d}) and could not provide a valid solution for \textit{lu}. The uniform configuration with $w=q=4096$ produced a similar speedup in 2 kernels (\textit{gemm} and \textit{correlation}), and a small speedup in \textit{lu}, failing to speed up \textit{correlation}. Despite the task requiring a longer budget than the one used for the accuracy benchmarks, this uniform configuration preserved its 410 token/s throughput until the end of the run, while the baseline model reached the 256k token limit at 140 token/s.

\subsection{Implementation and Computational Environment}
\subsubsection{Implementation in vLLM}

We implement the method inside vLLM~\cite{8}, a widely used inference engine. 
vLLM supports all six models in this study; however, none of them allow changing the attention of a full-attention layer through model configuration. We replace the attention class of each model at construction time, so that only the layers selected by a configuration are modified, without editing the weights or the configuration files. The replacement is installed as a vLLM plugin, so the method runs on an unmodified vLLM installation. Appendix~\ref{app:vllm} details the overrides, the Triton kernels, and how the configuration is verified on every GPU. Our implementation is publicly available at GitHub\footnote{https://github.com/jespb/CombinedSparseAttention}

\subsubsection{Computational environment }
Our experiments were run on two systems. The first is a node with 2x AMD EPYC 7453 28-Core, 503 GiB Host memory, 7x NVIDIA A100 80GB PCIe (1935 GB/s (HBM2e) and peak BF16 of 312 TFLOP/s), with vLLM (0.24.0), PyTorch (2.11.0+cu129), transformers (5.13.0), datasets (5.0.1), Python (3.12), attention backends flash\_attn (Qwen3.5) and triton\_attn (Gemma~4 and gpt-oss, whose head sizes and attention sinks require it), JIT-compiled by Triton 3.6.0, compilation: vllm\_compile (inductor), CUDA graph mode: full\_and\_piecewise. The second is a node with 4x NVIDIA GB200 (189 GB each, aarch64), used for Qwen3.5-122B-A10B-FP8 and Ornith-1.0-397B-FP8 with tensor parallelism of 4, with vLLM (0.27.1), PyTorch (2.13.0+cu130) and the flashinfer attention backend.



\subsection{Limitations}
\label{sec:limitations}

\par \textbf{Speedups only at long context:} Our configurations are slower than the baseline at short context, and only become faster after 32k to 64k tokens, depending on the model and the batch size. When the prompt is short, the first part of the output is therefore generated more slowly than by the unmodified model. The method is not a general solution to the slowdown of LLMs: it is a trade-off that pays off when long contexts dominate the workload.

\par \textbf{Memory:} Quest keeps the KV cache at full size (any page may be selected), so the method reduces compute but not memory. Plain SWA frees KV blocks on vLLM's sliding path and so also reduces memory, but at the cost of lowering the model's accuracy. Similarly, SnapKV also saves on memory by retaining only the top tokens of the input, but the static sinks did not perform well in preliminary work.

\section{Conclusions}

Hybrid models reduce the cost of long-context inference by replacing most of their attention layers with linear or sliding-window layers, but the full-attention layers that remain dominate the decoding time at long context. We linearize the computational cost of the full attention layers of 6 models by limiting the amount of tokens they attend to to the first 256 tokens, the last $W$ tokens (sliding window attention), and the $Q$ most relevant tokens using Quest. $W$ and $Q$ are two parameters that had to be discovered for each layer, resulting in a heterogeneous LLM configuration.

\par 
For each model, we compare 6 different configurations: the baseline model; two uniform configurations with the same parameters in all layers ($Q$=1024, $W$=256) that update the top tokens every step and every 32 steps; and three configurations obtained using NSGA-II as a multi-objective search algorithm (\textit{Best Accuracy, Best Speed, and Balanced}).

\par 
Our results show that we can obtain 1.16--1.26x and 2.06--2.23x speedups (batch size 4) at context lengths of 64k and 256k tokens, respectively, when using hybrid MoE models (Qwen3.5-122B and Ornith-397B), or dense models with a low parameter count (Qwen3.5-4B). For other dense models (Qwen3.5-27B and Gemma-4-31B), the speedups obtained were limited to 1.11x and 1.09x, respectively. For gpt-oss-120b, we only obtained minor speedups near the model's context length limit. The experiments showed no visible loss in the accuracy of the proposed configurations.

\par
We motivate this work by asking how much we can restrain the attention of full-attention layers without accuracy degradation. These results obtained by combining three sparse-attention methods tell us that we can place heavy restraints in the models, as long as the sparse-attention method is dynamic. 

\FloatBarrier

\bibliography{_bibliography}
\bibliographystyle{mlsys2026}


\FloatBarrier

\clearpage
\newpage

\appendix
\section{Appendix}

\subsection{Use case: Polybench}
\label{poly}

Despite the configurations successfully passing the long-context benchmarks, we try applying them to a simple use case: optimizing 4 polybench benchmarks. To do so, we select 4 kernels () and optimize all of them in parallel using batch size 4. For each kernel we make several prompts asking the Qwen3.5-4B model for additional optimization:

\par \textit{Prompt 1:} See Listing~\ref{lst:prompt1}, applied to gemm.
\par \textit{Prompts 2--4:} ``focus on memory access order: loop interchange so the innermost loop walks contiguous memory."
\par \textit{Prompts 5--8:} ``focus on cache blocking (tiling); experiment with the tile sizes."
\par \textit{Prompts 9--12:} ``parallelize with OpenMP over the 4 threads (work split, scheduling, no data races)."
\par \textit{Prompts 13--16:} ``focus on SIMD: make the inner loops vectorize (restrict, \#pragma omp simd, unrolling, AVX2/FMA intrinsics)."
\par \textit{Prompts 17--20:} ``combine the best ideas so far and tune their parameters."
\par \textit{Prompts 21+:} Until a context length of 256k tokens is reached, we alternate every 4 attempts between ``Try a substantially different loop structure or algorithmic formulation." and ``Tune your best version further: tile sizes, unroll factors, scheduling, alignment."

\begin{figure}[t]
\begin{lstlisting}[language=Python, caption={Prompt 1, applied to gemm.}, label={lst:prompt1}]
You are optimizing a C kernel from the PolyBench/C 4.2.1 suite for speed. The full benchmark file is below; only the function `kernel_gemm` is yours to change.

Machine and build:
- AMD EPYC 7453 (Zen 3): AVX2 + FMA, no AVX-512; 4 cores for your program (OMP_NUM_THREADS=4).
- gcc 13: `gcc -O3 -march=native -fopenmp -DEXTRALARGE_DATASET`. Problem sizes come from gemm.h, EXTRALARGE dataset.
- DATA_TYPE is double.

Rules:
- Keep the exact signature of `kernel_gemm`. You may add helper functions, `#include`s (e.g. <omp.h>, <immintrin.h>) and pragmas above it. No external libraries (no BLAS).
- The output arrays must match the original to a relative tolerance of 1e-06.

Each attempt is compiled, checked and timed (median of 3 runs) and I report the result. We will iterate for many attempts: keep improving on your best version.

Reply with the complete `kernel_gemm` function (and any helpers) in a single ```c code block.

gemm.h:
```c <full header>```

gemm.c:
```c <full source>```
\end{lstlisting}
\end{figure}

\begin{figure}[ht]
    \centering
    \includegraphics[width=\linewidth]{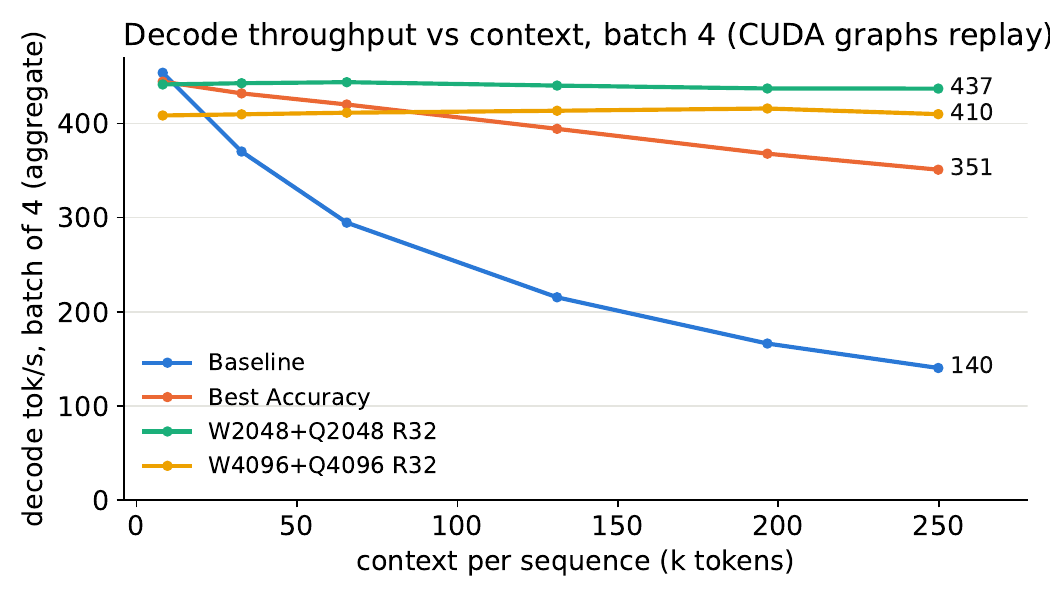}
    \caption{Throughput of the baseline Qwen3.5-4B model, best-accuracy configuration, and two uniform configurations with SWA and Quest budgets or 2048 and 4096 tokens.}
    \label{fig:pb_tps}
\end{figure}

\begin{figure*}
    \centering
    \includegraphics[width=\linewidth]{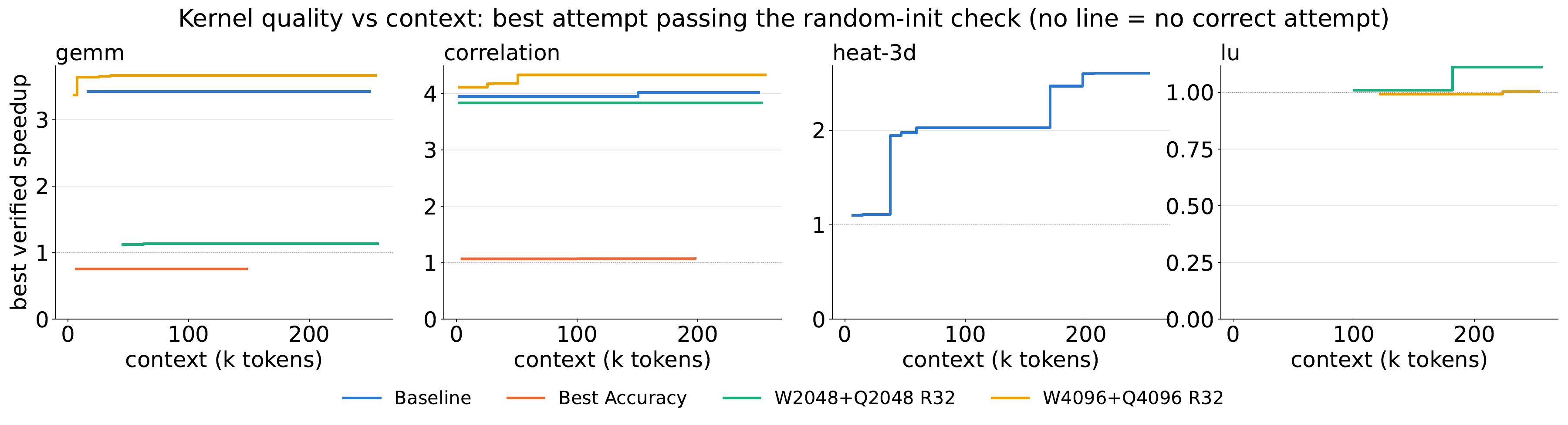}
    \caption{Speedup obtained in 4 polybench kernels (gemm, correlation, heat-3d and lu) as the conversation with each configuration moved towards 256k tokens. The lack of lines implies there was no correct code in that interval.}
    \label{fig:pb_quality}
\end{figure*}

In Figure \ref{fig:pb_tps} we see that the decode speed does not slow down even when moving long-context. Results wise, as seen in Figure \ref{fig:pb_quality}, despite this being a small non-coding model, it obtained a speedup in 3 out of 4 kernels, providing no correct solutions in the last kernel. By comparison, the uniform configuration also showed similar results, obtaining a speedup in 3 ou of 4 kernels. The \textit{best-accuracy} configuration, despite passing all the 21 long-context benchmarks, was unable to provide optimizations in this task.

\subsection{Modifications to vLLM}
\label{app:vllm}

\subsubsection{Per-layer sliding windows in vLLM}

As previously mentioned, our pipeline is based on vLLM~\cite{8}. 
vLLM supports per-layer windows, allowing us to easily replace full-attention layers with SWA, which changes both the KV-cache allocation and the arguments passed to the FlashAttention kernel. Since the window values are defined per layer, it allows us to easily explore using different window sizes over the model, rather than experimenting with a homogeneous SWA implementation.

\par \textbf{Qwen3.5:} vLLM's implementation reuses Qwen3NextAttention, never passing a window, and the window is not reachable through any config override. Instead of editing the config, we use the prototype in Listing~\ref{lst:sliding-window-attention} as a wrapper, allowing us to define a window size.

\begin{figure}
\begin{lstlisting}[language=Python, caption={Sliding-window attention override}, label={lst:sliding-window-attention}]
def attention_with_sliding_window(*args, prefix: str = "", **kwargs):
    window = windows.get(extract_layer_index(prefix))
    if window is not None:
        kwargs["per_layer_sliding_window"] = window
    return original_attention(*args, prefix=prefix, **kwargs)

qwen3_next.Attention = attention_with_sliding_window
\end{lstlisting}
\end{figure}

This window is applied to prompt and generation jointly: a query at position t attends to [t-W+1, t], with no attention sink or global tokens retained. This is the plain banded mask of Longformer's local component, not StreamingLLM's sink-preserving variant~\cite{13}.

\subsubsection{Query-aware sparse attention in vLLM}

Our pipeline is based on vLLM~\cite{8}. In none of the six models is the attention of a full-attention layer reachable through a config override: Qwen3.5 reuses Qwen3NextAttention, and Gemma~4 and gpt-oss build their layers from the module-level \texttt{Attention} class. Instead of editing the configs or the weights, we replace that class in each model module with the factory in Listing~\ref{lst:attention-factory}. It reads the layer index from the module prefix vLLM builds while constructing the model and rewires only the layers selected by the configuration.

\begin{figure}[!t]
\begin{lstlisting}[language=Python, caption={Per-layer attention override}, label={lst:attention-factory}]
def attention_factory(*args, prefix: str = "", **kwargs):
    layer = original_attention(*args, prefix=prefix, **kwargs)
    idx = extract_layer_index(prefix)
    if idx in config:  
        window, budget = config[idx]
        apply_quest_attention(layer, window, budget, pinned=num_sinks)
    return layer

for module in(qwen3_next, gemma4, gpt_oss):
    module.Attention = attention_factory
\end{lstlisting}
\end{figure}

\par \textbf{Learned sinks:} For gpt-oss, the per-head sink logit $s_h$ is added to the softmax denominator of the sparse attention, $\sum_{j} e^{q\cdot k_j - m} + e^{s_h - m}$, which reproduces the model's attention exactly when every page is selected.

\par \textbf{Multi-GPU deployment:} With tensor parallelism, vLLM builds the model in spawned worker processes, which do not inherit the override in the parent process. A vLLM plugin reinstalls it in every worker from an environment variable. After initialisation, every rank reports which layers it modified, and the run is aborted if any rank differs from the intended configuration.

\FloatBarrier

\subsection{Profilling vLLM}

\subsubsection{Full Attention vs Uniform Configurations}

\begin{figure*}
    \centering
    \includegraphics[width=0.95\linewidth]{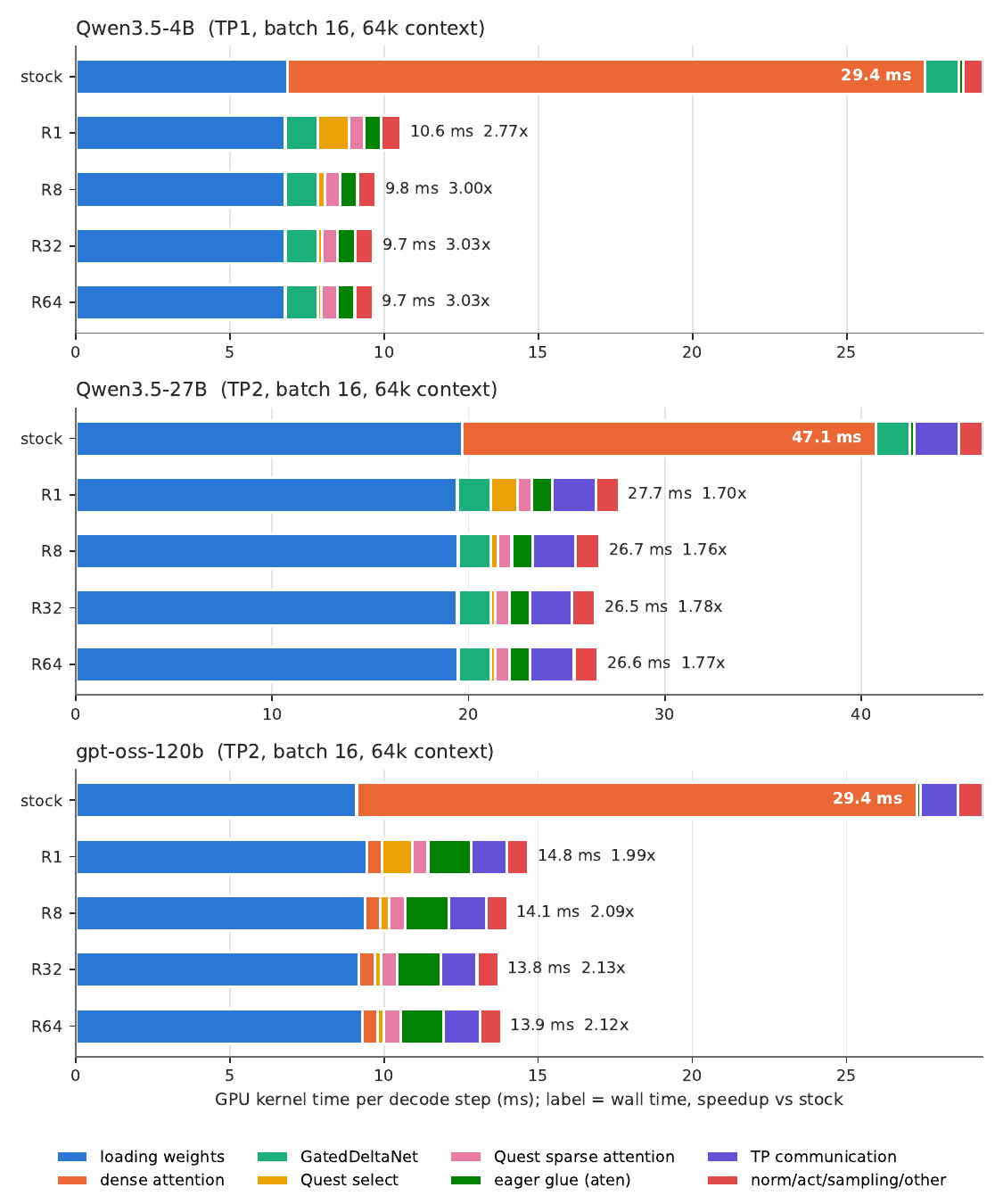}
    \caption{Average time spent in each kernel category when decoding 256 tokens, starting at a context length of 64k tokens.}
    \label{fig:profile_decode}
\end{figure*}

This section gives illustrates the time spent in each kernel category when decoding long-context sequences. Figure \ref{fig:profile_decode} illustrates the average time spent decoding each token when the context length is already at 64k tokens. We see that despite replacing full attention with quest saves us a lot in compute, the next bottleneck of the system is the time spent in loading weights. We also see that updating Quest at every decode step results in the throughput being slower when compared to decoding every 8, 32 or 64 steps. Despite we updating Quest every 32 tokens, this figure shows that there would be little loss in throughput if we updated every 8 tokens instead.

\subsection{Throughput}

\begin{figure*}
  \centering
  Single-GPU setup: A100\\
  \begin{minipage}{0.46\textwidth}
      \centering
      \includegraphics[width=\textwidth,trim=0 25 0 0,clip]{images/throughput_Qwen3_5-4B_1GPU.pdf}
  \end{minipage}
  \hfill
  \begin{minipage}{0.46\textwidth}
      \centering
      \includegraphics[width=\textwidth,trim=0 25 0 0,clip]{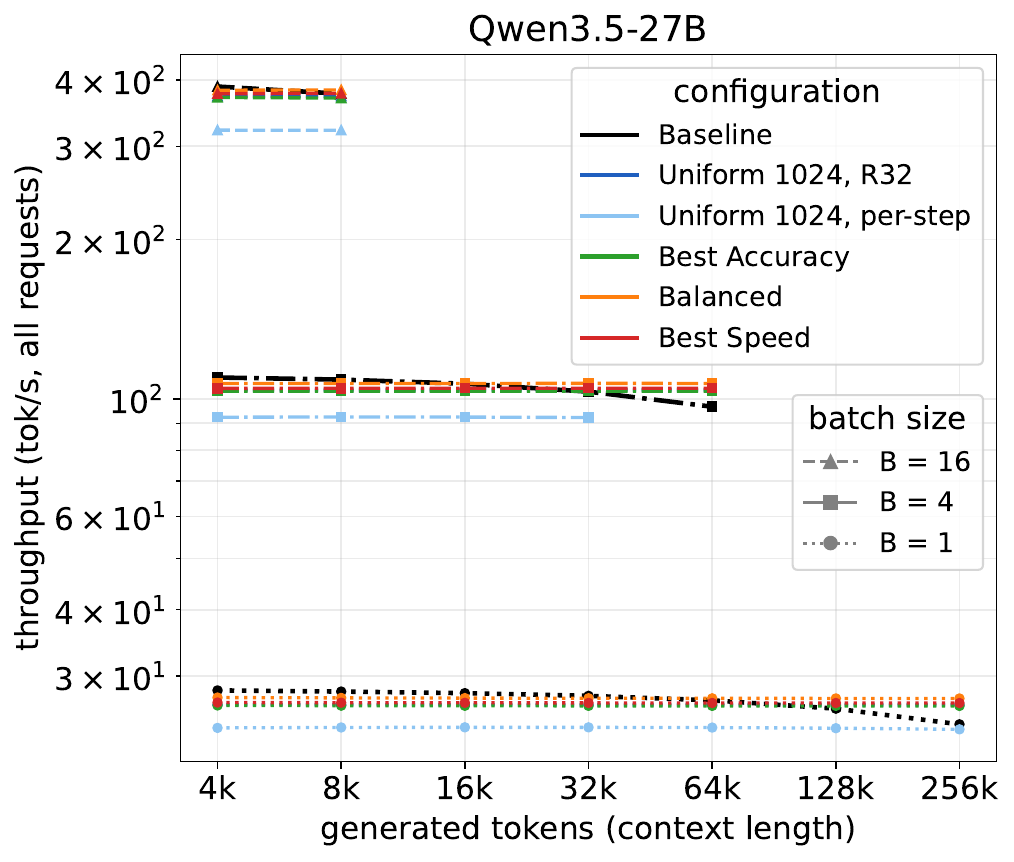}
  \end{minipage}

  \vspace{0.5em}

  Dual-GPU setup: A100s\\
  \begin{minipage}{0.46\textwidth}
      \centering
      \includegraphics[width=\textwidth,trim=0 25 0 0,clip]{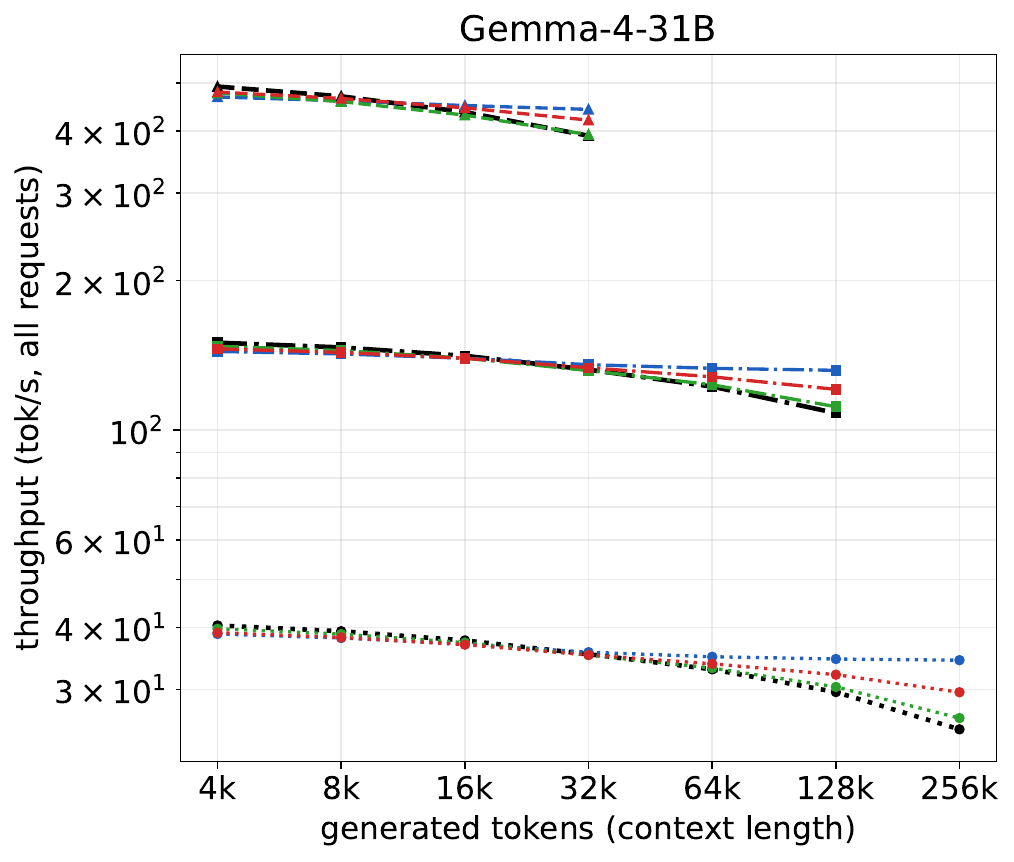}
  \end{minipage}
  \hfill
  \begin{minipage}{0.46\textwidth}
      \centering
      \includegraphics[width=\textwidth,trim=0 25 0 0,clip]{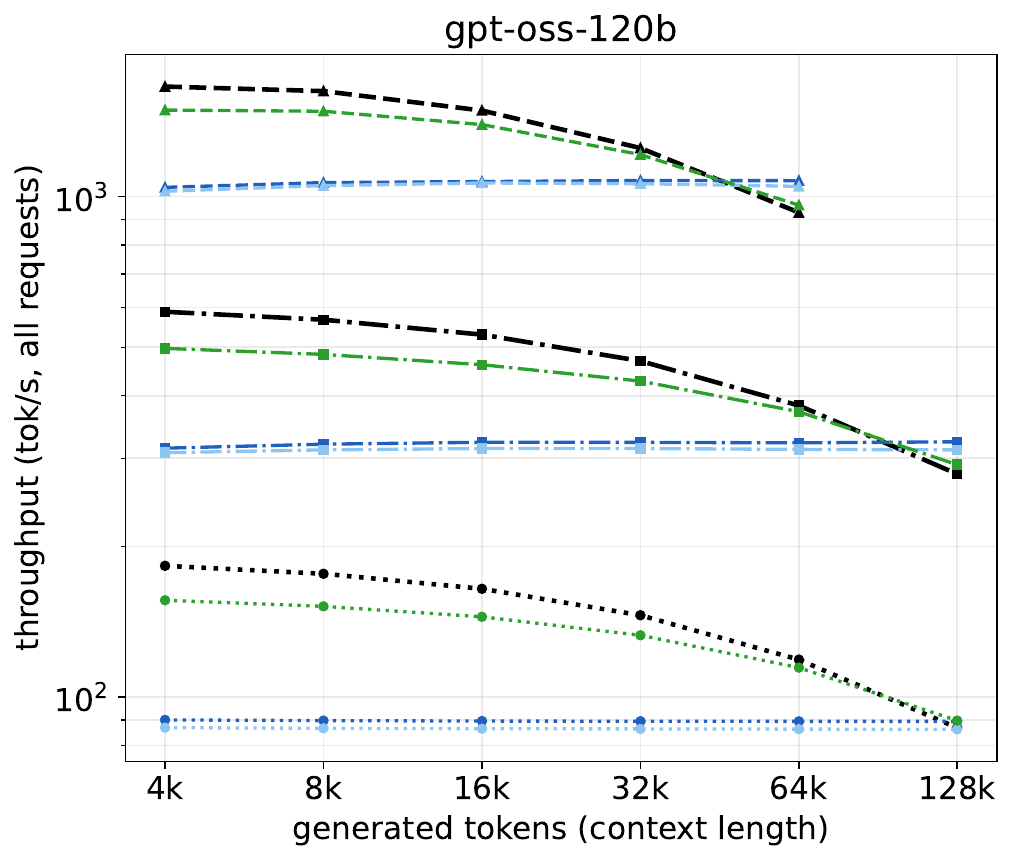}
  \end{minipage}

  \vspace{0.5em}

  Quad-GPU setup: GB200 node\\
  \begin{minipage}{0.46\textwidth}
      \centering
      \includegraphics[width=\textwidth]{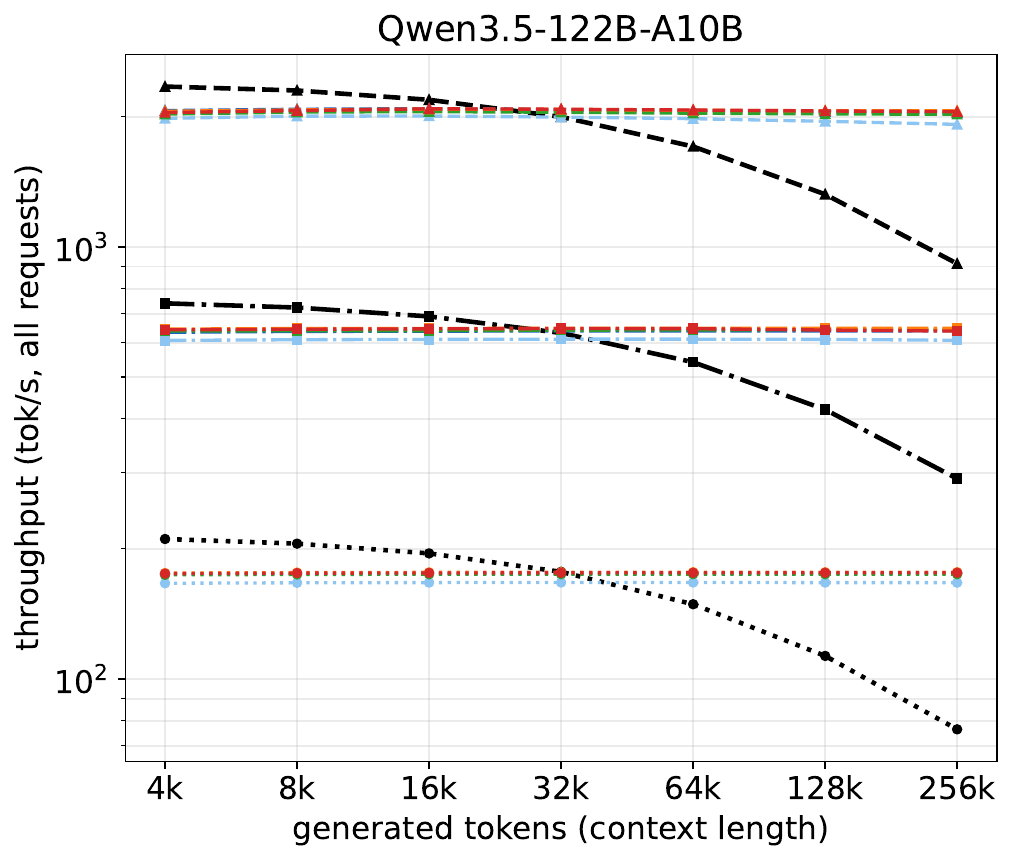}
  \end{minipage}
  \hfill
  \begin{minipage}{0.46\textwidth}
      \centering
      \includegraphics[width=\textwidth]{images/throughput_Ornith-397B_4GPU.pdf}
  \end{minipage}

  \caption{Tokens-per-second throughput using different configurations and batch sizes.
  We see that the baseline models are faster for short sequences due to the computation overhead from Quest. For longer sequences starting from 64k tokens, we see reasonable speedups in Qwen-4B, Qwen-122B, and Ornith. These speedups range 1.16--1.26x in context lengths of 64k tokens and 2.06--2.23x for 256k tokens}
  \label{fig:throughput}
  \end{figure*}

Figure~\ref{fig:throughput} shows the tokens per second throughput of all configurations using difference batch sizes.

\subsection{Configurations}
\label{app:configs}
Table~\ref{tab:configs} shows the configurations obtained for each model.

\FloatBarrier

\begin{table*}[!b]
    \centering
    \vspace{-25mm}
    \caption{Configurations evaluated for each model: the window $W$ and Quest budget $B$ (tokens) of every full-attention layer, indexed by its position in the model. `full' marks a layer left at full attention, whose budget is irrelevant (`--'). On gpt-oss-120b, the search front contained a single configuration; on Ornith, Best Speed is the Uniform 1024 configuration.}
    \scriptsize
    \setlength{\tabcolsep}{2pt}
    \begin{tabular}{l l cccccccc}
    \toprule
    \multicolumn{2}{l}{\textbf{Qwen3.5-4B}} & 3 & 7 & 11 & 15 & 19 & 23 & 27 & 31 \\
    \midrule
    Best Accuracy & $W$ & 256 & 4096 & 1024 & 4096 & full & 2048 & 256 & 1024 \\
     & $B$ & 512 & 256 & 512 & 2048 & -- & 2048 & 2048 & 2048 \\
    \midrule
    Balanced & $W$ & 256 & 512 & 256 & 4096 & 256 & 256 & 256 & 256 \\
     & $B$ & 1024 & 256 & 512 & 2048 & 1024 & 512 & 512 & 2048 \\
    \midrule
    Best Speed & $W$ & 256 & 256 & 256 & 256 & 256 & 256 & 256 & 256 \\
     & $B$ & 512 & 512 & 512 & 512 & 512 & 512 & 512 & 512 \\
    \midrule
    Uniform 1024 & $W$ & 256 & 256 & 256 & 256 & 256 & 256 & 256 & 256 \\
     & $B$ & 1024 & 1024 & 1024 & 1024 & 1024 & 1024 & 1024 & 1024 \\
    \bottomrule
    \end{tabular}
    \par\medskip
    \begin{tabular}{l l cccccccccccccccc}
    \toprule
    \multicolumn{2}{l}{\textbf{Qwen3.5-27B}} & 3 & 7 & 11 & 15 & 19 & 23 & 27 & 31 & 35 & 39 & 43 & 47 & 51 & 55 & 59 & 63 \\
    \midrule
    Best Accuracy & $W$ & 4096 & 256 & 256 & 256 & 256 & 256 & 256 & 2048 & 256 & 256 & 256 & 256 & 4096 & 2048 & 4096 & 256 \\
     & $B$ & 1024 & 512 & 512 & 256 & 1024 & 512 & 512 & 512 & 512 & 4096 & 1024 & 256 & 512 & 512 & 512 & 512 \\
    \midrule
    Balanced & $W$ & 256 & 256 & 256 & 256 & 256 & 256 & 256 & 256 & 256 & 256 & 256 & 256 & 256 & 256 & 256 & 256 \\
     & $B$ & 512 & 2048 & 512 & 512 & 512 & 512 & 512 & 512 & 512 & 512 & 512 & 512 & 512 & 512 & 512 & 512 \\
    \midrule
    Best Speed & $W$ & 256 & 256 & 256 & 256 & 256 & 256 & 256 & 256 & 256 & 256 & 256 & 256 & 256 & 256 & 256 & 256 \\
     & $B$ & 512 & 512 & 512 & 512 & 512 & 512 & 512 & 512 & 512 & 512 & 512 & 512 & 512 & 512 & 512 & 512 \\
    \midrule
    Uniform 1024 & $W$ & 256 & 256 & 256 & 256 & 256 & 256 & 256 & 256 & 256 & 256 & 256 & 256 & 256 & 256 & 256 & 256 \\
     & $B$ & 1024 & 1024 & 1024 & 1024 & 1024 & 1024 & 1024 & 1024 & 1024 & 1024 & 1024 & 1024 & 1024 & 1024 & 1024 & 1024 \\
    \bottomrule
    \end{tabular}
    \par\medskip
    \begin{tabular}{l l cccccccccc}
    \toprule
    \multicolumn{2}{l}{\textbf{Gemma-4-31B}} & 5 & 11 & 17 & 23 & 29 & 35 & 41 & 47 & 53 & 59 \\
    \midrule
    Best Accuracy & $W$ & full & 1024 & full & full & full & full & full & full & full & 4096 \\
     & $B$ & -- & 1024 & -- & -- & -- & -- & -- & -- & -- & 1024 \\
    \midrule
    Best Speed & $W$ & full & 4096 & 512 & 1024 & 256 & full & full & full & 1024 & 2048 \\
     & $B$ & -- & 256 & 256 & 1024 & 1024 & -- & -- & -- & 2048 & 2048 \\
    \midrule
    Uniform 1024 & $W$ & 256 & 256 & 256 & 256 & 256 & 256 & 256 & 256 & 256 & 256 \\
     & $B$ & 1024 & 1024 & 1024 & 1024 & 1024 & 1024 & 1024 & 1024 & 1024 & 1024 \\
    \bottomrule
    \end{tabular}
    \par\medskip
    \begin{tabular}{l l cccccccccccccccccc}
    \toprule
    \multicolumn{2}{l}{\textbf{gpt-oss-120b}} & 1 & 3 & 5 & 7 & 9 & 11 & 13 & 15 & 17 & 19 & 21 & 23 & 25 & 27 & 29 & 31 & 33 & 35 \\
    \midrule
    Best Acc. / Speed & $W$ & full & 4096 & 1024 & full & full & full & full & full & full & full & full & full & 256 & full & full & full & 512 & full \\
     & $B$ & -- & 256 & 1024 & -- & -- & -- & -- & -- & -- & -- & -- & -- & 512 & -- & -- & -- & 1024 & -- \\
    \midrule
    Uniform 1024 & $W$ & 256 & 256 & 256 & 256 & 256 & 256 & 256 & 256 & 256 & 256 & 256 & 256 & 256 & 256 & 256 & 256 & 256 & 256 \\
     & $B$ & 1024 & 1024 & 1024 & 1024 & 1024 & 1024 & 1024 & 1024 & 1024 & 1024 & 1024 & 1024 & 1024 & 1024 & 1024 & 1024 & 1024 & 1024 \\
    \bottomrule
    \end{tabular}
    \par\medskip
    \begin{tabular}{l l cccccccccccc}
    \toprule
    \multicolumn{2}{l}{\textbf{Qwen3.5-122B-A10B-FP8}} & 3 & 7 & 11 & 15 & 19 & 23 & 27 & 31 & 35 & 39 & 43 & 47 \\
    \midrule
    Best Accuracy & $W$ & 256 & 256 & 256 & 256 & 256 & 256 & 256 & 256 & 256 & 256 & 256 & 512 \\
     & $B$ & 1024 & 1024 & 1024 & 2048 & 1024 & 1024 & 1024 & 1024 & 1024 & 4096 & 1024 & 1024 \\
    \midrule
    Balanced & $W$ & 256 & 256 & 256 & 256 & 256 & 256 & 256 & 512 & 256 & 256 & 256 & 512 \\
     & $B$ & 1024 & 1024 & 1024 & 1024 & 1024 & 1024 & 1024 & 1024 & 1024 & 1024 & 1024 & 512 \\
    \midrule
    Best Speed & $W$ & 256 & 256 & 256 & 256 & 256 & 256 & 256 & 256 & 256 & 256 & 256 & 512 \\
     & $B$ & 1024 & 1024 & 1024 & 1024 & 1024 & 512 & 1024 & 1024 & 1024 & 1024 & 1024 & 1024 \\
    \midrule
    Uniform 1024 & $W$ & 256 & 256 & 256 & 256 & 256 & 256 & 256 & 256 & 256 & 256 & 256 & 256 \\
     & $B$ & 1024 & 1024 & 1024 & 1024 & 1024 & 1024 & 1024 & 1024 & 1024 & 1024 & 1024 & 1024 \\
    \bottomrule
    \end{tabular}
    \par\medskip
    \begin{tabular}{l l ccccccccccccccc}
    \toprule
    \multicolumn{2}{l}{\textbf{Ornith-1.0-397B-FP8}} & 3 & 7 & 11 & 15 & 19 & 23 & 27 & 31 & 35 & 39 & 43 & 47 & 51 & 55 & 59 \\
    \midrule
    Best Accuracy & $W$ & 256 & 4096 & 256 & 2048 & 2048 & 256 & 256 & full & 512 & 512 & 256 & 512 & full & 4096 & 256 \\
     & $B$ & 1024 & 4096 & 256 & 4096 & 512 & 1024 & 4096 & -- & 4096 & 1024 & 1024 & 2048 & -- & 1024 & 1024 \\
    \midrule
    Balanced & $W$ & 512 & 256 & 256 & 256 & 256 & 256 & 256 & 256 & 256 & 256 & 256 & 256 & 256 & 256 & 1024 \\
     & $B$ & 1024 & 256 & 1024 & 4096 & 1024 & 1024 & 1024 & 1024 & 1024 & 1024 & 1024 & 1024 & 1024 & 1024 & 1024 \\
    \midrule
    Best Speed / Uniform 1024 & $W$ & 256 & 256 & 256 & 256 & 256 & 256 & 256 & 256 & 256 & 256 & 256 & 256 & 256 & 256 & 256 \\
     & $B$ & 1024 & 1024 & 1024 & 1024 & 1024 & 1024 & 1024 & 1024 & 1024 & 1024 & 1024 & 1024 & 1024 & 1024 & 1024 \\
    \midrule
    Uniform 256 & $W$ & 256 & 256 & 256 & 256 & 256 & 256 & 256 & 256 & 256 & 256 & 256 & 256 & 256 & 256 & 256 \\
     & $B$ & 256 & 256 & 256 & 256 & 256 & 256 & 256 & 256 & 256 & 256 & 256 & 256 & 256 & 256 & 256 \\
    \bottomrule
    \end{tabular}
    \label{tab:configs}
\end{table*}

\end{document}